\documentclass[10pt,twocolumn,letterpaper]{article}

\usepackage{cvpr}
\usepackage{times}
\usepackage{epsfig}
\usepackage{graphicx}
\usepackage{amsmath}
\usepackage{amssymb}

\usepackage{gensymb}
\usepackage{subcaption}
\usepackage{xcolor}

\usepackage[breaklinks=true,colorlinks,bookmarks=false]{hyperref}

\cvprfinalcopy 

\def\cvprPaperID{****} 

\begin{document}

\title{Lending Orientation to Neural Networks for Cross-view Geo-localization}


\author{Liu Liu $^{1,2}$ and Hongdong Li $^{1,2}$\\
$^{1}$ Australian National University, Canberra, Australia \\ 
$^{2}$ Australian Centre for Robotic Vision \\
\tt\small{ \{Liu.Liu; hongdong.li\}}@anu.edu.au
}

\maketitle

\begin{abstract}
This paper studies image-based geo-localization (IBL) problem using ground-to-aerial cross-view matching.  The goal is to predict the spatial location of a ground-level query image by matching it to a large geotagged aerial image database (\eg, satellite imagery).   This is a challenging task due to the drastic differences in their viewpoints and visual appearances. Existing deep learning methods for this problem have been focused on maximizing feature similarity between spatially close-by image pairs, while minimizing other images pairs which are far apart. They do so by deep feature embedding based on visual appearance in those ground-and-aerial images.  However, in everyday life, humans commonly use {\em orientation} information as an important cue for the task of spatial localization.  Inspired by this insight, this paper proposes a novel method which endows deep neural networks with the `commonsense' of orientation.  Given a ground-level spherical panoramic image as query input (and a large georeferenced satellite image database), we design a Siamese network which explicitly encodes the orientation (\ie, spherical directions) of each pixel of the images. Our method significantly boosts the discriminative power of the learned deep features, leading to a much higher recall and precision outperforming all previous methods.  Our network is also more compact using only 1/5th number of parameters than a previously best-performing network.  To evaluate the generalization of our method, we also created a large-scale cross-view localization benchmark containing 100K geotagged ground-aerial pairs covering a city. Our codes and datasets are available at \url{https://github.com/Liumouliu/OriCNN}.
\end{abstract}

\vspace{-0.2in}
\begin{figure}
\centering

\begin{subfigure}[t]{0.15\textwidth}
\includegraphics[width=\textwidth]{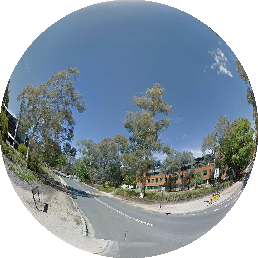}~~~
{\caption{\small{Query image}}}
\end{subfigure}
\begin{subfigure}[t]{0.30\textwidth}
\includegraphics[width=\textwidth]{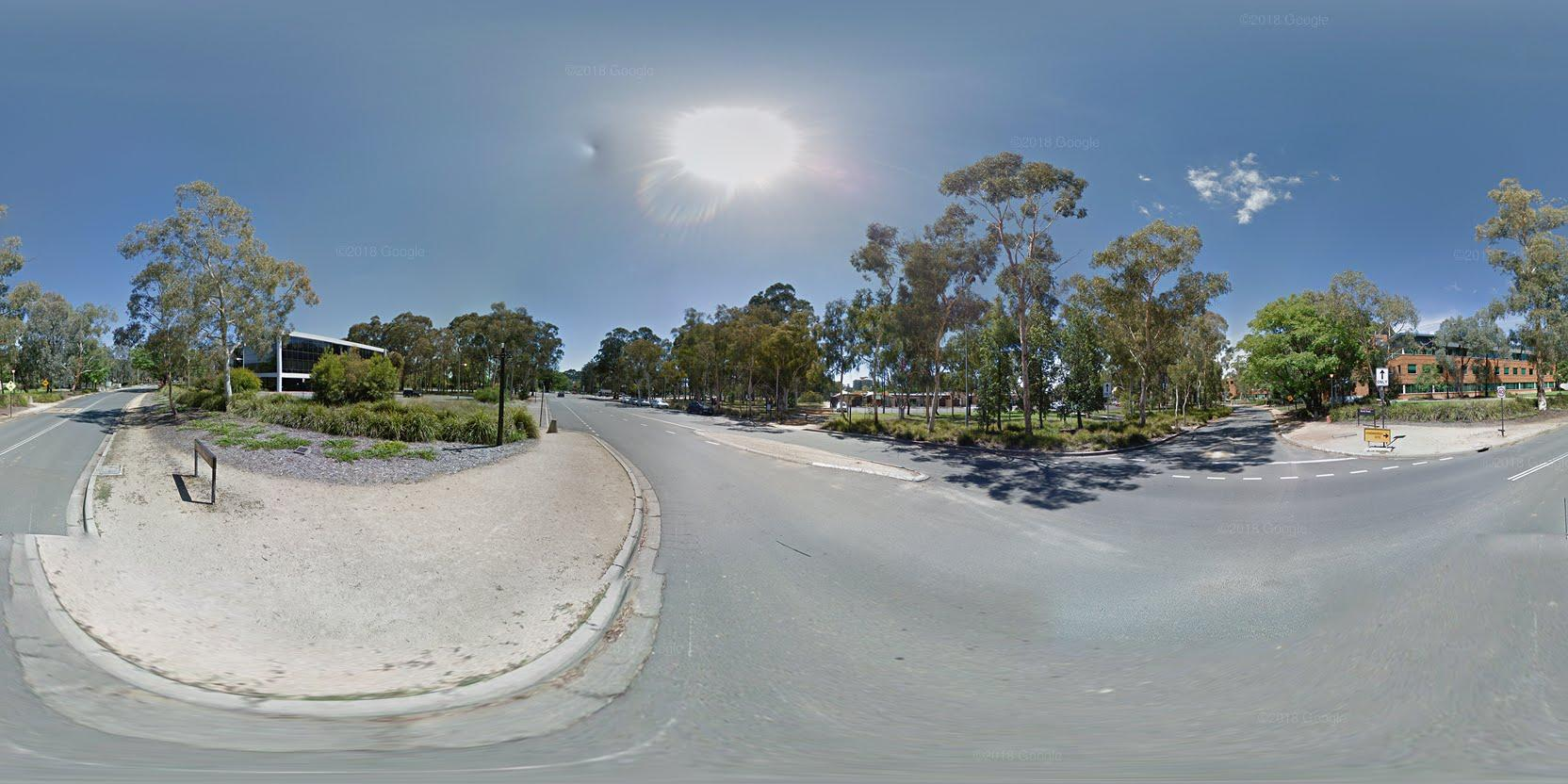}
{\caption{\small{Ground-level panorama}}}
\end{subfigure}
\begin{subfigure}[t]{0.253\textwidth}
\includegraphics[width=\textwidth]{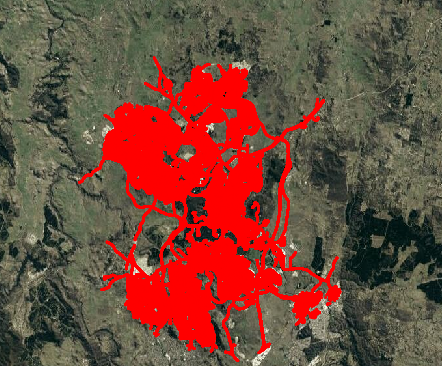}
\caption{\small{Our satellite dataset (with GPS footprints)}}
\end{subfigure}~~~~~~~
\begin{subfigure}[t]{0.15\textwidth}
\includegraphics[width=\textwidth]{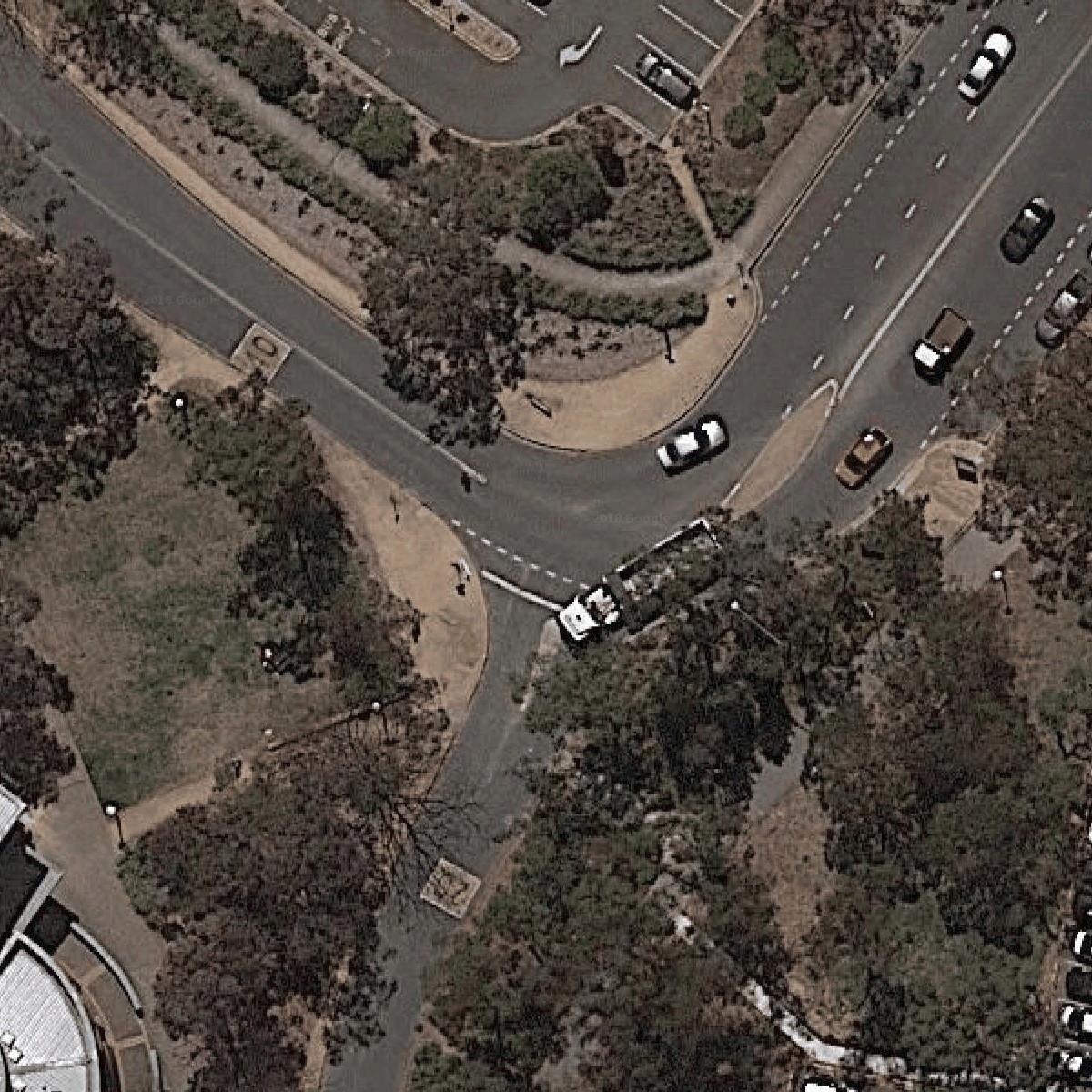}
\caption{\small{The matched satellite image}}
\end{subfigure}

\caption{\small {Given a ground-level spherical omnidirectional image (a) (or its panoramic representation as shown in (b)) as a query image, the task of image-based localization (IBL) is to estimate the location of the query image by matching it to a large satellite image database covering the same region, as given in (c).  The found correct match is shown in (d), which is centered at the very same location as (a).}
\label{fig:sampleImagePairANUdata}}
\end{figure}

\section{Introduction}
This paper investigates the problem of image-based geo-localization using ground-to-aerial image matching.  Given a ground-level query image, we aim to recover the absolute geospatial location at which the image is taken. This is done by comparing the query image with a large collection of geo-referenced aerial images (\eg, satellite images) without the aid of other localization sensors (such as GPS).  Figure \ref{fig:sampleImagePairANUdata} illustrates an example scenario of such ground-to-aerial cross-view localization problem.

While previous image-based geolocalization methods have been primarily based on ground-to-ground (street-view level) image matching (\eg, \cite{arandjelovic2016netvlad,radenovic2018fine,DBLP:journals/corr/abs-1808-08779,Liu_2017_ICCV,liu2018robust,sattler2019understanding}), using ground-to-aerial cross-view matching (\ie, matching ground-level photos to aerial imagery ) is becoming an attractive approach for image-based localization, thanks to the widespread coverage (over the entire Earth) and easy accessibility of satellite and aerial survey images.  In contrast, the coverage of ground-level street views (such as Google-map or Bing-map) is at best limited to urban areas. 

Another advantage of using ground-to-aerial matching, which has been overlooked in recent IBL literature, is that the way to localize using ground-to-aerial matching resembles what a human would localize himself using a traditional paper map.  A map can be considered as a (coarse) aerial image depicting the geographic region. Imagine the following scenario where a tourist is lost in a foreign city, yet he has no modern localization aid with him (\eg, no GPS, no cell phone, no google map) except for a paper-version tourist map. In such circumstance, a natural way for him to re-localize (by himself) is to match the city map (\ie, an aerial drawing) with what he sees (\ie, a ground-level street view). In this human localization scenario,  to re-orientate himself (\ie, knowing where the geographic {\em True North} is, both on the map and in the surroundings) is critically important,  which will greatly simplify the localization task. 

Inspired by this insight,  we propose a novel deep convolutional neural network that explicitly encodes and exploits orientation (directional) information aiming for more accurate ground-to-aerial localization.  Specifically, we intend to teach a deep neural-network the concept of ``direction" or ``orientation" at every pixel of the query or database images. We do so by creating an orientation map, used as additional signal channels for the CNN. We hope to learn more expressive feature representations which are not only appearance-driven and location-discriminative, but also orientation-selective.  The novel way we propose to incorporate orientation information is compact, thus our orientation map can be easily plugged to other deep-learning frameworks or for other applications as well.

Our work makes the following contributions:
{ (1) A simple yet efficient way to incorporate per-pixel orientation information to CNN for cross-view localization; (2)  A novel Siamese CNN architecture that jointly learns feature embeddings from both appearance and orientation geometry information; (3) Our method establishes a new state-of-the-art performance for ground-to-aerial geolocalization; Besides, as a by-product of this work, we also created a new large-scale cross-view image benchmark dataset (CVACT) consisting of densely covered and geotagged street-view panoramas and satellite images covering a city.  CVACT is much bigger than the most popular CVUSA \cite{zhai2017predicting} panorama benchmark (Ref. Table-\ref{dataset_compare}). We hope this will be a valuable addition and contribution to the field.} 
\section{Related Works}
\noindent\textbf{Deep cross-view localization.} 
Traditional hand-crafted features were used for cross-view localization \cite{lin2013cross,castaldo2015semantic,DBLP:journals/corr/MousavianK16}.
With the success of modern deep learning based methods, almost all state of the art cross-view localization methods (\cite{Hu_2018_CVPR,workman2015location,workman2015wide,vo2016localizing}) adopted deep CNN to learn discriminative features. Workman \etal \cite{workman2015location} demonstrated that deep features learned from a classification model pre-trained on Places \cite{zhou2014learning} dataset outperforms those hand-crafted features.  They further extended the work to cross-view training and showed improved performance \cite{workman2015wide}. Vo \etal \cite{vo2016localizing} explored several deep CNN architectures (\eg, Classification, Hybrid, Siamese and Triplet CNN) for cross-view matching, and proposed a soft margin triplet loss to boost the performance of triplet embedding. They also give a deep regression network to estimate the orientation of ground-view image and utilize multiple possible orientations of aerial images to estimate the ground-view orientation. This causes significant overhead in both training and testing. Instead, we directly incorporate the cross-view orientations to CNN to learn discriminative features for localization. Recently, Hu \etal \cite{hu2018cvm} proposed to use NetVLAD \cite{arandjelovic2016netvlad} as feature aggregation on top of the VGG \cite{simonyan2014very} architecture pre-trained on ImageNet dataset \cite{deng2009imagenet}, and obtained the state-of-the-art performance. Paper \cite{lin2015learning,tian2017cross} tackled the ground to bird-eye view (45-degree oblique views) matching problem, which is a relatively easier task in the sense that more overlaps (\eg, building facade) between a ground-level and a 45-degree view can be found.

\noindent\textbf{Transfer between ground and aerial images.}
The relationship between a ground-view image and a satellite image is complex; no simple geometric transformation or photometric mapping can be found other than very coarse homography approximation or color tune transfer.  Existing attempts \cite{zhai2017predicting,regmi2018cross} used a deep CNN to transfer a ground-view panorama to an aerial one, or vice versa, in various approaches including the conditional generative adversarial networks \cite{regmi2018cross}.  Zhai \etal \cite{zhai2017predicting}  proposed  to  predict  the  semantic of a ground-view image from its corresponding satellite image. The predicted semantic is used to synthesize ground-view panorama. 

\noindent\textbf{Lending orientation to neural networks.}  There was little work in the literature addressing lending directional information to neural networks, with a few exceptions. An early work by Zamel \etal \cite{zemel1995lending} introduced the idea of using complex-valued  neurons to represent (one-dimensional) direction information.  Their idea is to equip  neural networks with directional information based on the observation that a directional signal in the 2D plane can be coded as a complex number, $z= |z|\exp{(-i\phi)}$, of which the phase component  naturally represents the angular directional variable $\phi~$.  By this method, conventional real-valued neural networks can be extended to complex-valued neural networks, for which all the learning rules (such as back-propagation can be adapted accordingly, and directional information is handled in a straightforward way.  Some recent works further developed this idea \cite{guberman2016complex,trabelsi2018deep}.   

\section{Method Overview} 

\subsection{Siamese Network Architecture}

Using ground-to-aerial cross-view matching for image-based localization is a challenging task.  This is mainly because of the ultra-wide baseline between ground and aerial images, which causes a vast difference in their visual appearances despite depicting the same geographic region.  Such difficulty renders conventional local descriptor based method (\eg, Bag of SIFTs) virtually useless.  Deep-learning has been adopted for solving this problem, and obtained remarkable success (\eg, \cite{Hu_2018_CVPR,vo2016localizing,workman2015location,workman2015wide,lin2015learning}).  A common paradigm that has been used by those deep methods is to formulate the problem as {\em feature embedding}, and extract location-sensitive features by training a deep convolutional neural network. They do so by forcefully pulling positive ground-and-aerial matching pairs to be closer in feature embedding space, while pushing those features coming from non-matchable pairs far apart.  

\begin{figure}[!h]
\begin{center}
\includegraphics[width=0.35\textwidth]{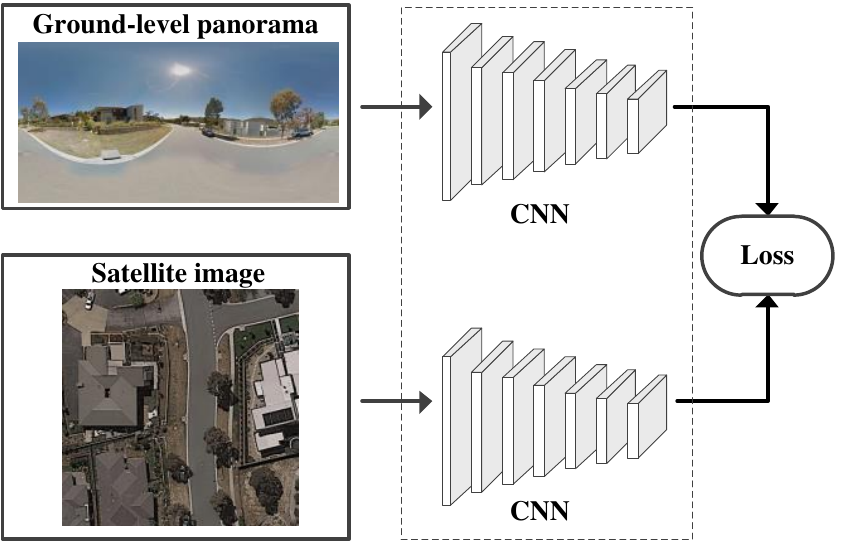}
\end{center}
\caption{\small{Our baseline Siamese network: the inputs to the two branches are ground-level panoramas and satellite images, respectively. Features are learned by minimizing a triplet loss \cite{Hu_2018_CVPR}.}}
\label{fig:Siamesenet}
\end{figure}

In this paper, we adopt a Siamese-type two-branch CNN network of 7 layers (showing in Figure \ref{fig:Siamesenet}) as the basis of this work.  Each branch learns deep features that are suitable for matching between the two views.  Unlike previous Siamese networks for ground-to-ground localization, the two branches of our Siamese net do not share weights, because the domains (and modalities) of ground and aerial imagery are different. It is beneficial to allow more degree of freedom for their own weights to evolve. 

\subsection{Use of Orientation Information}
As discussed previously, we notice that most of those previous deep localization networks all focus on capturing image similarity in terms of visual appearance (and semantic content); they have all overlooked an important geometric cue for localization, namely, the orientation or directional information of the views - an important cue that humans (and many animals) often use for spatial awareness and localization \cite{Animal_navigation}. 

On the other hand, knowing the orientation (\ie, knowing in which direction every point in an image is pointing to) will greatly simplify the localization task.  Almost all off-the-shelf satellite image databases are georeferenced, in which the geographic `North' direction is always labeled on the images.  In the context of image-based localization, if one is able to identify the True North on a ground-level query image, then that image can be placed in a geometrically meaningful coordinate frame relative to the satellite images' reference frame.  This will significantly reduce the search space for finding the correct ground-to-satellite matches, resulting in much faster searches and more accurate matches. To show the importance of knowing orientation with respect to the task of localization, let us return to our previous example, and examine what a disoriented tourist would do in order to quickly relocalize himself in a foreign city with a paper map. First, he needs to identify in which direction lies the geographic True North in the foreign street he is standing in;  Second, face the True North direction, and at the same time rotate the paper map so that its `North' is pointing the same direction;  Third, look in certain directions and try to find some real landmarks (along those directions) that match the landmarks printed on the map. Finding enough good matches suggests that the location of him is recovered.  

This paper provides an efficient way to teach (and to endow with) deep neural networks the notion of (geographic) orientation -- this is one of the key contributions of the work.  Next, we explain how we encode such per-pixel orientation information in a convolutional neural network.  

\subsection {Representing Orientation Information}  
Many real-world problems involve orientation or directional variables, \eg, in target detection with radar or sonar, microphone array for stereo sound/acoustic signal processing, and in robot navigation.  However, few neural networks addressed or exploited such directional information, with only a few exceptions including the complex-valued Boltzmann network \cite{zemel1995lending}. While our method to be described in this paper was originally inspired by paper \cite{zemel1995lending},  we find it unsuitable for the task of image-based localization, for two reasons.  First, in image-based localization, the direction of each pixel is actually two dimensional (\ie, the two spherical angles parameterized by ($(\theta, \phi)$) --azimuth and altitude), rather than a single phase angle.  There is no simple way to represent two angles with a single complex number.  Second, both the time and memory complexities of a complex-valued network are expensive. Given these reasons, we abandoned the idea of using a complex network. Instead, we propose a very simple (and straightforward) way to directly inject per-pixel orientation information via {\em orientation maps}. Details are to be given next. 

\paragraph{Parameterization.}
\begin{figure}[!h]
\begin{center}
\includegraphics[width=0.48\textwidth]{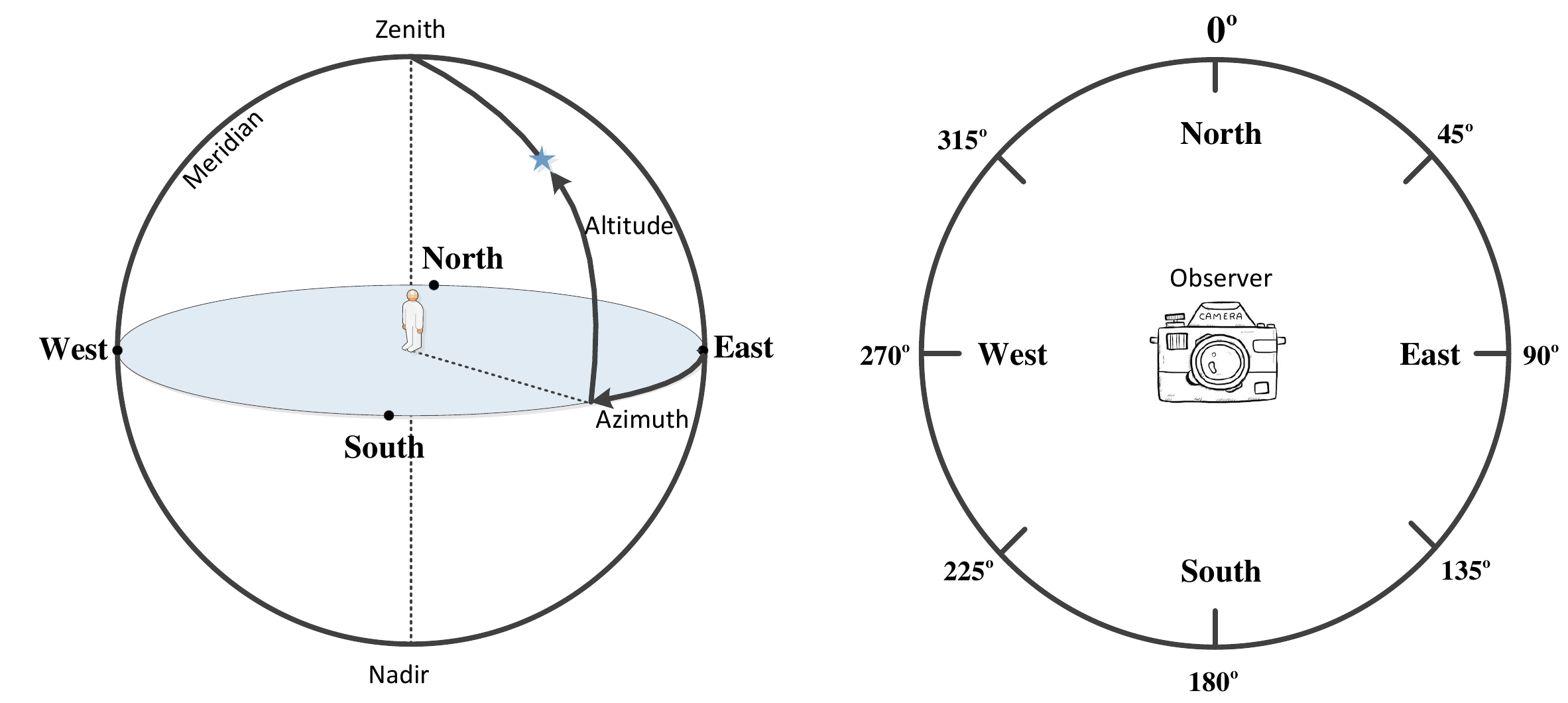}
\end{center}
   \caption{\small{We use spherical angles (azimuth and altitude) to define the orientation at each pixel of a ground-level panorama (shown in the left), and use polar coordinates (azimuth and range) to define the orientation for pixels in a satellite image (shown in the right).}}
\label{fig:orientation-definition}
\end{figure}

We consider a ground-level query image as a spherical view covering a full omnidirectional $360\degree\times 180\degree$ Field-of-View (FoV).  The orientation of each pixel therein is parameterized by two spherical angles: azimuth and altitude $\theta$ and $\phi$.  The mapping between a spherical image to a rectangular panoramic image can be done by using {\em equirectangular projection}. Note that, in order to know the relative angle between pixels we assume the input panorama image is intrinsically calibration. Getting intrinsic calibration is an easy task. Moreover, for the sake of IBL, even a very coarse estimate of the camera intrinsics is adequate for the task.  Since satellite view captures an image orthogonal to the ground plane, without loss of generality, we assume the observer is standing at the center location of the satellite view.  We then simply use polar angle (in the polar coordinate system) to represent the azimuth angle $\theta_i$, and range $r_i$ to represent the radial distance of a pixel in the satellite image relative to the center, \ie, $r_i =(y_i^2+x_i^2)^{1/2};~~\theta_i=\arctan2(y_i,x_i)$.

\begin{figure}[!h]
\begin{center}
\includegraphics[width=0.25\textwidth, height=0.11\textwidth,]{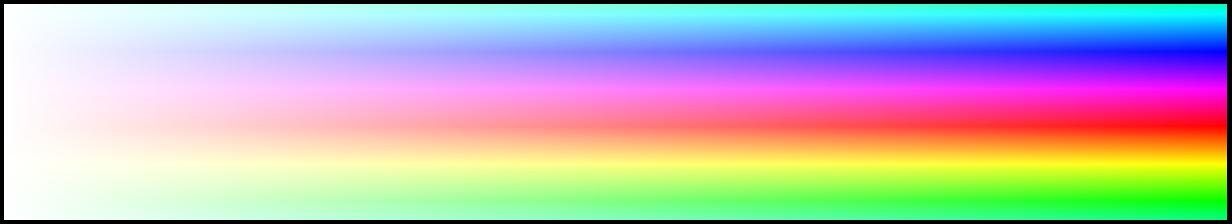}~~~~
\includegraphics[width=0.12\textwidth]{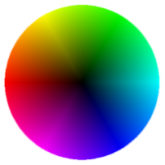}
\end{center}
\caption{\small{Color-coded orientation maps (\ie, U-V maps) . Left: U-V map for ground-level panorama; Right: U-V map for aerial view.}}
\label{fig:colormapUV}
\end{figure}

\paragraph{Color-coded orientation maps.} 
We borrow the same color-coding scheme developed for visualizing 2D optical-flow field \cite{Toolbox_optical} to represent our 2D orientation map.  Specifically, the hue (H) and saturation (S) channels in a color map each represents one of the two orientation parameters. Specifically, for a ground-level panorama, the two channels are $\theta$ (azimuth) and $\phi$ (altitude), and for an overhead satellite image, the two channels are $\theta$ (azimuth) and $r$ (range). This way, we can simply consider the orientation map is nothing but two additional color-channels (we denote them as U-V channels), besides the original 3-channel RGB input image.  
Figure \ref{fig:colormapUV} shows such two color orientation maps.

\section{Joint Image and Orientation Embedding} \label{sec::Embedding}
Now that with the above preparations in place, we are ready to describe our joint image and orientation feature embedding method. 
\subsection{Where to inject orientation information?}
Our network is based on the Siamese architecture of $7$ convolutional layers, which are cropped from the generator network in view synthesis \cite{regmi2018cross,isola2017image}. Each layer consists of convolution, leaky-ReLU, and batch-normalization. Implementation details are deferred to Section-\ref{sec::implementation}.

\begin{figure}[!h]
\begin{center}
\includegraphics[width=0.5\textwidth]{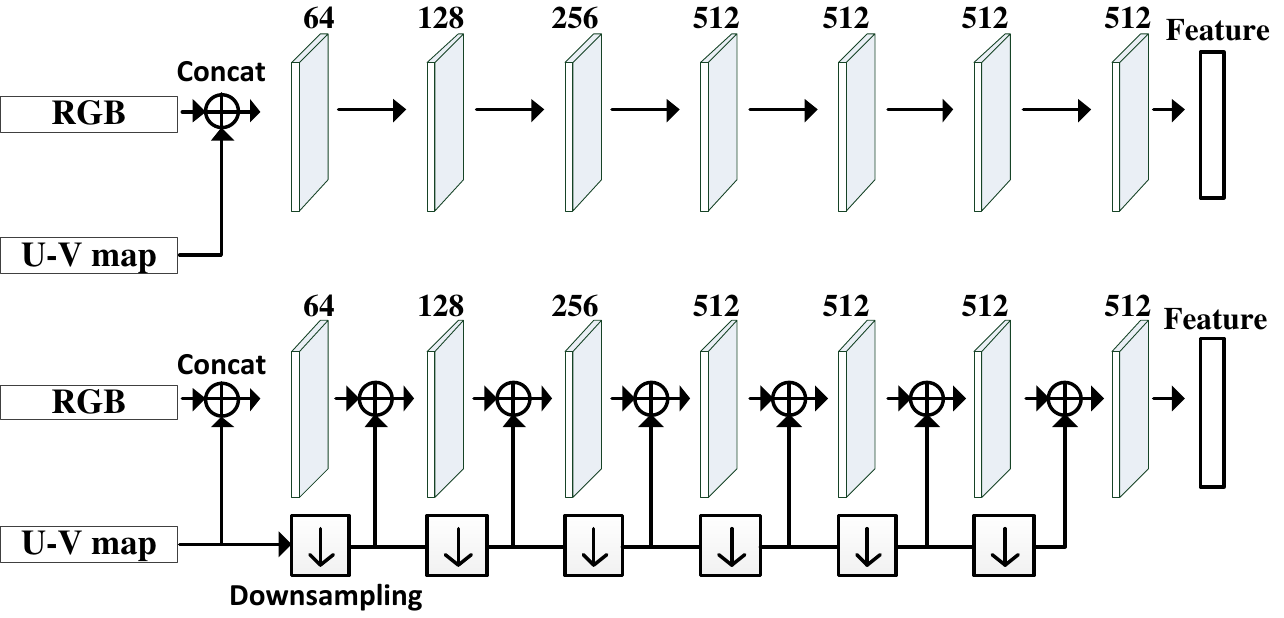}
\caption{\small{Two schemes to incorporate orientation information. Scheme-I (top): orientation information (U-V map) are injected to the input layer only; Scheme-II (bottom): orientation information (U-V map) are injected to all layers.}}
\label{fig:cnnArt2}
\end{center}
\end{figure}

We devise two different schemes (Scheme-I, and Scheme-II) for injecting orientation information to the Siamese net at different layers.  In Scheme-I, we concatenate cross-view images and orientation masks along RGB channels to yield cross-view inputs. In Scheme-II, besides concatenating cross-view images and orientation masks as inputs, we also inject orientation information to intermediate convolutional blocks. Down-sampled cross-view orientation maps are concatenated with output feature map of each convolutional block. These two schemes are illustrated in Figure \ref{fig:cnnArt2}. 

\subsection{Deep feature embedding}

\begin{figure}[tp]
\begin{center}
\includegraphics[width=0.45\textwidth]{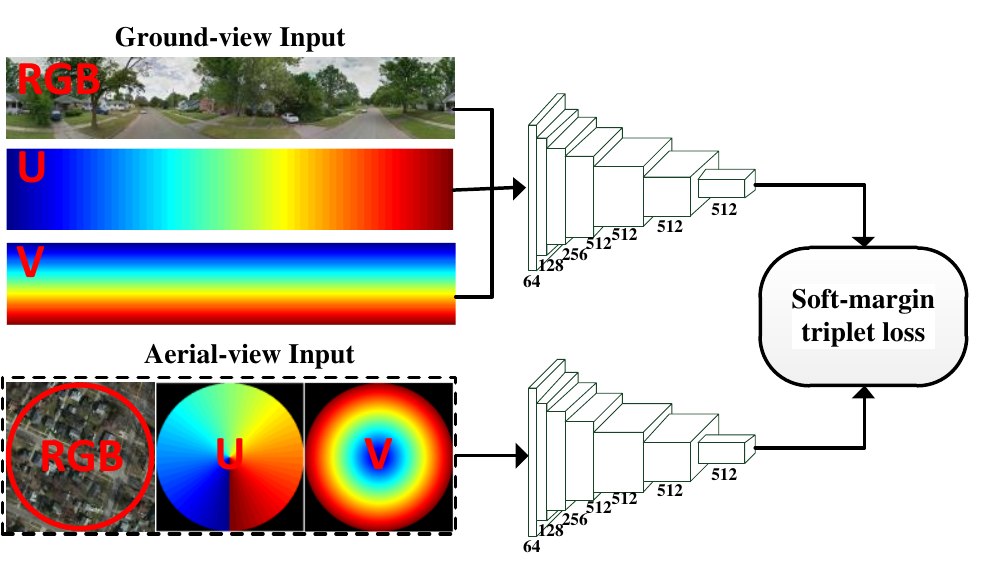}
\end{center}
   \caption{\small{Our overall network architecture (in \textit{Scheme-I}). Cross-view images and their associated orientation maps are jointly fed to the Siamese net for feature embedding. The learned two feature vectors are passed to a triplet loss function to drive the network training. The numbers next to each layer denote the number of filters.}}
\label{fig:cnnArt}
\end{figure}
We aggregate feature maps obtained at the last three layers to form multi-scale feature vectors. Intermediate feature maps are resized and concatenated along the feature dimension to form a 3D tensor $\mathcal{X}_i, i\in \{g,s\}$ of $W_i$x$H_i$x$D$ dimension, where $W_i$, $H_i$, and $D$ are the width, height, and feature dimension, respectively.



We aim to extract a compact embedding from $\mathcal{X}_i$. We add a pooling layer acting on $\mathcal{X}_i$ and outputs a vector $\mathbf{f}_i$. Since cross-view images usually have different sizes, we adopted the generalized-mean pooling (proposed in \cite{radenovic2018fine,dollar2009integral}) to get the following embedding: {
$\mathbf{f}_i = \left[f_i^1,..,f_i^{D} \right]^T, f_i^k=\left ( \frac{1}{W_iH_i}\sum_{w=1}^{W_i}\sum_{h=1}^{H_i}x_{w,h,k}^p \right )^{1/p}$}. Variable $x_{w,h,k}$ is a scalar at the $k$-th feature map of tensor $\mathcal{X}_i$, and $p$ is a scalar. We set $D=1536$, and normalize all $\mathbf{f}_i'$s to be unit $L_2$-norm.

\subsection{Triplet loss for cross-view metric learning} \label{sec::Learning}
Given embeddings $\mathbf{f}_g$ and $\mathbf{f}_s$ of the ground-view panorama and satellite image, respectively, 
the aim of cross-view metric learning is to embed the cross-view embeddings to a same space, with metric distances ($L_2$-metric) between embeddings reflect the similarity/dissimilarity between cross-view images. There are many metric learning objective functions available, \eg, triplet ranking \cite{arandjelovic2016netvlad}, SARE \cite{DBLP:journals/corr/abs-1808-08779}, contrastive \cite{radenovic2018fine}, angular \cite{Wang_2017_ICCV} losses. All losses try to pull the $L_2$ distances between matchable cross-view embeddings, while pushing the $L_2$ distances among non-matchable cross-view embeddings. We adopt the weighted soft-margin ranking loss \cite{Hu_2018_CVPR} to train our Siamese net for its state-of-the-art performance in this cross-view localization task. The loss function $\mathcal{L}$ is defined by:
$\mathcal{L} = \log\left \{1+\exp\left [ \alpha\left ( \left \| \mathbf{f}_{g} - \mathbf{f}_{s} \right \|^2 - \left \| \mathbf{f}_{g} - \mathbf{f}_{s}^* \right \|^2 \right ) \right ] \right \}$, where $\mathbf{f}_{g}$ and $\mathbf{f}_{s}$ are features from matchable cross-view pair, and $\mathbf{f}_{s}^*$ is non-matchable to $\mathbf{f}_{g}$. $\alpha$ is a parameter chosen empirically.

\section{Experiments} \label{sec::Experiments}
\label{sec::implementation}

\begin{figure}[!ht]
\begin{center}
\includegraphics[width=0.35\textwidth, height=0.1\textwidth]{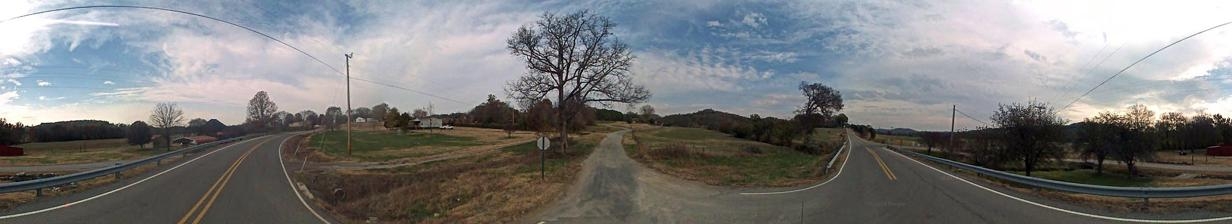}~
\includegraphics[width=0.1\textwidth]{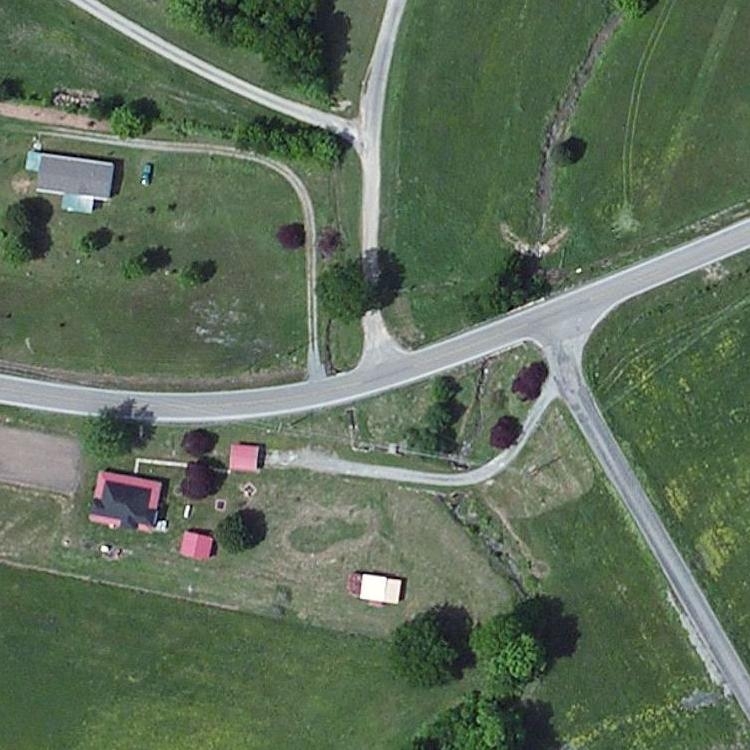}
\end{center}
\caption{\small{A sample ground-level panorama and satellite image pair from CVUSA dataset. 
}}
\label{fig:semantic_seg}
\end{figure}

\paragraph{Training and testing datasets.} 
We use the CVUSA panorama dataset \cite{zhai2017predicting} to evaluate our method.  CVUSA is a standard cross-view dataset, containing 35,532 ground-and-satellite image pairs for training, and 8,884 for testing.  It has been popularly used by many previous works (\cite{Hu_2018_CVPR,workman2015wide, vo2016localizing, zhai2017predicting}), thus allows for easy benchmarking.   
A sample pair of a ground-level panorama and satellite image from CVUSA is displayed in Figure \ref{fig:semantic_seg}. 
In the course of this research, in order to evaluate our network's generalization ability, we also collected and created a new (and much larger) cross-view localization benchmark dataset -- which we call the {\em CVACT dataset} -- containing 92,802 testing pairs (\ie, 10$\times$ more than CVUSA) with ground-truth geo-locations.  Details about the CVACT dataset will be given in a later subsection, Section-\ref{Sec::ACTdataset}. 

\vspace{3pt}
\noindent\textbf{Implementation details.} We train our 7-layer Siamese network from scratch, \ie, from randomly initialized network weights (with zero mean and $0.02$ stdv. Gaussian noise).  We use $4\times4$ convolution kernel throughout and strides at $2$ with zero padding.  The smaller slope for the of Leaky-ReLU is $0.2$. Momentum in batch normalization is set at $0.1$, and gamma is randomly initialized with a Gaussian noise with unit mean, and stdv=$0.02$. In computing the triplet loss, we use the same $\alpha=10$ as in \cite{Hu_2018_CVPR}.  For the generalized-mean pooling layer, we set $p = 3$ as recommended by \cite{radenovic2018fine}.  Our CNN is implemented in Tensorflow using Adam optimizer \cite{kingma2014adam} with a learning rate of $10^{-5}$ and batch size of $B=12$.  We use exhaustive mini-batch strategy \cite{vo2016localizing} to maximize the number of triplets within each batch. Cross-view pairs are fed to the Siamese net. For each ground-view panorama, there are $1$ positive satellite image and $B-1$ negative satellite images, resulting in total $B(B-1)$ triplets. For each satellite image, there are also $1$ positive ground-view panorama and $B-1$ negative ground-view panoramas, resulting in total $B(B-1)$ triplets. In total, we employ $2B(B-1)$ triplets.  

\vspace{3pt}
\noindent\textbf{Data augmentation.} To improve our network's robustness against errors in the global orientation estimation of a query image, we adopt `data augmentation' strategy. This is done by circularly shifting the input ground-level panorama by a random relative angle, resulting in a random rotation of the ground-level image along the azimuth direction (\ie, estimated `True North' direction). 


\vspace{3pt}
\noindent\textbf{Evaluation metrics.} 
The most commonly used metric for evaluating the performance of IBL methods is the recall rates (among the found top-K candidates, where K = 1, 2, 3,...).  For ease of comparison with previous methods, we use {\em recalls} at top $1\%$ as suggested by \cite{Hu_2018_CVPR,vo2016localizing,workman2015wide} -- detailed definition can be found therein.  In this paper, we only display the recalls at top-1, top-5, top-10, up to top 1\%.


\subsection{Effect of Orientation Map}
This is our very first (and also very important) set of experiments, through which we intend to show that lending orientation information to deep network greatly improves the performance of cross-view matching based geo-localization. 

Recall that we have developed two different schemes of adding orientation information to our Siamese network. In Scheme-1 we simply augment the input signal from a 3-channel RGB image to having 5 channels (\ie, RGB + UV); and in Scheme-2 we inject the UV map to each of the seven CNN layers.  Our experiments however found no major difference in the performances by these two schemes. For this reason, in all our later experiments, we only test Scheme-1.  Scheme-1 is not only easy to use, but also can be plugged to any type of network architectures (\eg, VGG \cite{simonyan2014very}, ResNet \cite{he2016deep}, U-net \cite{ronneberger2015u}, or DenseNet \cite{huang2017densely}) without effort. Figure \ref{fig:cnnArt} gives our CNN architecture based on Scheme-1.

\vspace{3pt}
\noindent\textbf{Baseline network.} We first implemented a simple 7-layer Siamese net, and the net is trained using standard 3-channel RGB input. This is our baseline network for comparison. Note that all ground-view panoramas are aligned to the north direction. The first row of Table-\ref{tab::RGBEncoderWithAndWithoutMask} shows this baseline performance, namely, recalls for the top-1, top-5 and top-10  and top-1\% candidates are 9.8\%, 23.6\%, 32.6\%, and 68.6\%, respectively.  

\vspace{3pt}
\noindent\textbf{Our new network.} We then trained and tested our new method with 5-channel input (for both Scheme-1 and Scheme-2), and obtained much higher recalls throughout the experiments. The results are shown in the $2^{nd}$ and $3^{rd}$ rows of Table-\ref{tab::RGBEncoderWithAndWithoutMask}.  For example, by our Scheme-1 we obtained recalls for top-1, top-5, top-10, and top-1\% at 31.7\%, 56.6\%, 67.5\%, and 93.1\% respectively -- showing significant improvements of more than 25 percentage all-round.  

\begin{table}[]
\setlength{\tabcolsep}{4pt}
\caption{Recall performance on CVUSA dataset \cite{zhai2017predicting}.}
\begin{tabular}{|l|c|c|c|c|}
\hline
Method$|$Recalls  & r@1 & r@5 & r@10 & r@ top 1\%\\ \hline
Baseline (RGB)       &    9.83&     23.66&     32.68 &   68.61     \\ \hline
Our -I (RGB-UV)              & \textbf{31.71}          & 56.61          & 67.57  & \textbf{93.19}    \\ \hline
Our -II (RGB-UV)           & 31.46          & \textbf{57.22}          & \textbf{67.90}  & 93.15    \\ \hline


\end{tabular}
\label{tab::RGBEncoderWithAndWithoutMask}
\end{table}

\begin{figure}[t]
\begin{center}
\includegraphics[width=0.39\textwidth]{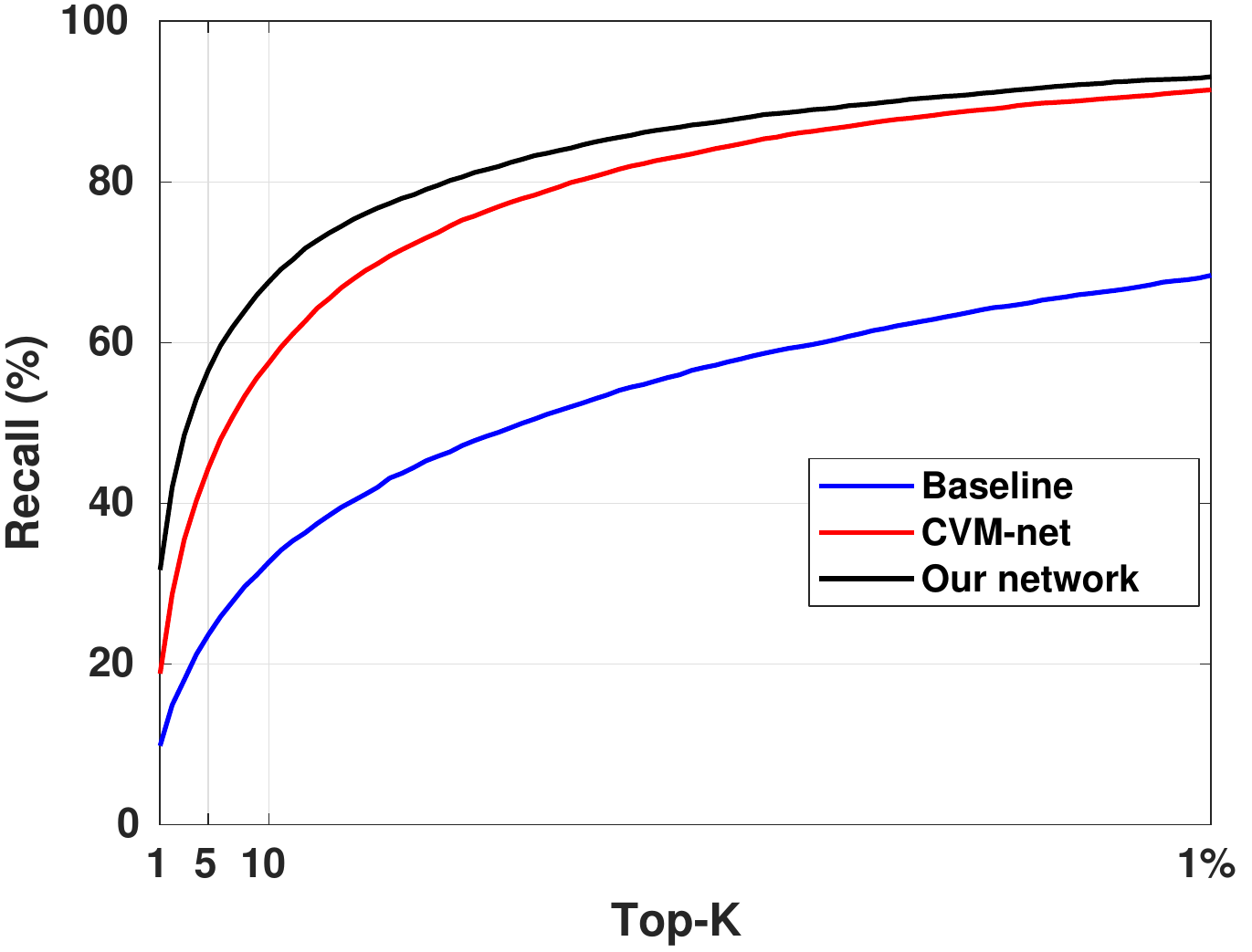}
\end{center}
   \caption{\small{This graph shows that, simply by exploiting orientation information to a baseline Siamese network (via U-V maps) we are able to boost the success rates (measured in recalls) by over 25\%. Our new method also outperforms the SOTA deep cross-view localization method of CVM-net.}}
\label{fig:recall_CVUSA}
\end{figure}

\subsection{Comparisons with Other Methods}
We compare our method with state-of-the-art methods, which include Workman \etal \cite{workman2015wide}, Vo \etal \cite{vo2016localizing}, Zhai \etal \cite{zhai2017predicting} and CVM-Net \cite{Hu_2018_CVPR}.   The results (of recall\@ top 1\%) are given in Table-\ref{tab::top1_per}. 

Table-\ref{tab::NetVLADCNN} gives more results in terms of recalls. We observe that 1)  CVM-net leverages the feature maps obtained by VGG16 net \cite{simonyan2014very} pre-trained on ImageNet \cite{deng2009imagenet}, and uses NetVLAD \cite{arandjelovic2016netvlad} for feature aggregation ; 2) Compared with CVM-net \cite{Hu_2018_CVPR}, our method achieves a relative improvement of $+1.65\%$ for recall@top $1\% $ and $+12.91\%$ for recall@top-1; 3) Our network is more compact than CVM-net, can be quickly trained from scratch. The total number of trainable parameters and storage cost of our net is $30$-millions and $368MB$, while in the case of CVM-Net \cite{Hu_2018_CVPR} the corresponding numbers are $160$-millions and $2.0GB$, respectively. Based on a single GTX1080Ti GPU the total training time for our 7-layer Siamese net took about 3 days on CVUSA dataset.  The average query time is only at 30 ms per query, about $1/3$ of CVM-net. We also experimented plugging our orientation map to a 16-layer VGG net, and similar improvements are obtained.

\begin{table}[]{\small{ 
\setlength{\tabcolsep}{2.5pt}
\caption{\small{Comparison of recall @top 1\% recalls by state-of-the-art methods on CVUSA dataset \cite{zhai2017predicting}.}}
\begin{tabular}{|c|c|c|c|c|c|}
\hline & Ours & Workman \cite{workman2015wide} & Zhai \cite{zhai2017predicting} & Vo \cite{vo2016localizing} & CVM-net \cite{Hu_2018_CVPR}  \\ \hline
r@top 1\% & \textbf{93.19}   & 34.30 & 43.20   & 63.70 & 91.54  \\ \hline
\end{tabular}
\label{tab::top1_per}}} 
\end{table}

\begin{table}[]{\small
\caption{\small Comparison of recall performance with CVM-net \cite{Hu_2018_CVPR} on CVUSA dataset \cite{zhai2017predicting}.}
\begin{tabular}{|l|c|c|c|c|}
\hline
Method                     & r@1            & r@5            & r@10  & r@top 1\%          \\ \hline
Our 7-layer network                         & \textbf{31.71}          & \textbf{56.61}          & \textbf{67.57}      & 93.19    \\ \hline

Our 16-layer VGG   & 27.15          & 54.66          & 67.54   & \textbf{93.91}       \\ \hline CVM-net \cite{Hu_2018_CVPR}      & 18.80          & 44.42          & 57.47   & 91.54       \\ \hline
\end{tabular}
\label{tab::NetVLADCNN}}
\end{table}
\subsection{Detailed analyses of the proposed network}
\begin{figure}[!ht]
\begin{center}
\includegraphics[width=0.35\textwidth]{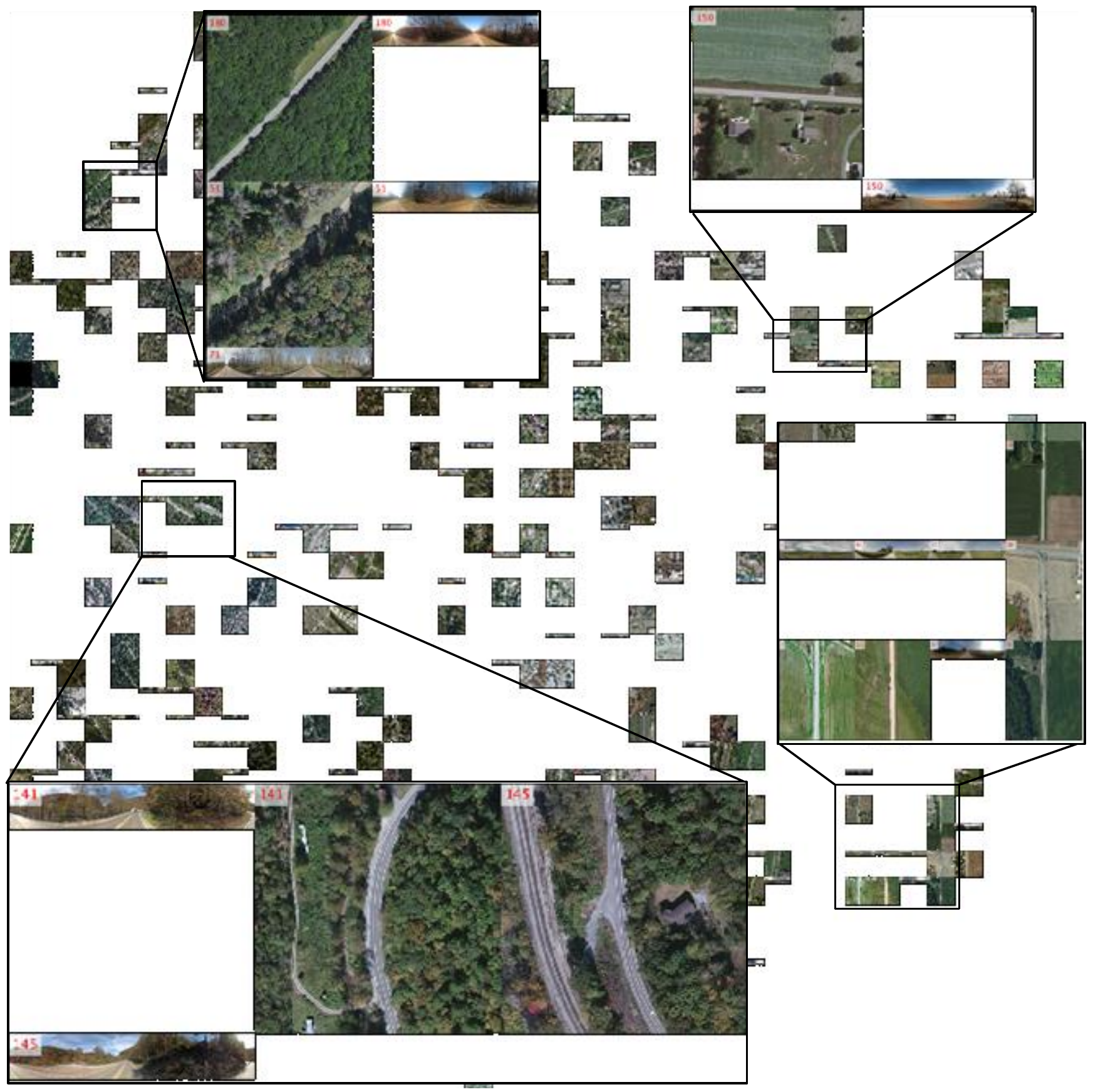}
\end{center}
   \caption{\small{t-SNE visualization of cross-view features learned by our method. The ID on the top-left corner of each image denotes the index of the cross-view pair \cite{maaten2008visualizing}. (Best viewed on screen with zoom-in)}}
\label{fig:tsne}
\end{figure}

\noindent\textbf{t-SNE visualization of the feature embedding.} Our network learns location-discriminative feature embeddings.  To visualize the embeddings, we plot those learned features in 2D using t-SNE \cite{maaten2008visualizing}.  Figure \ref{fig:tsne} shows a result for CVUSA \cite{zhai2017predicting}. Clearly, spatially near-by cross-view image pairs are embedded to close to each other.

\begin{figure}[t]
\begin{center}
\includegraphics[width=0.35\textwidth]{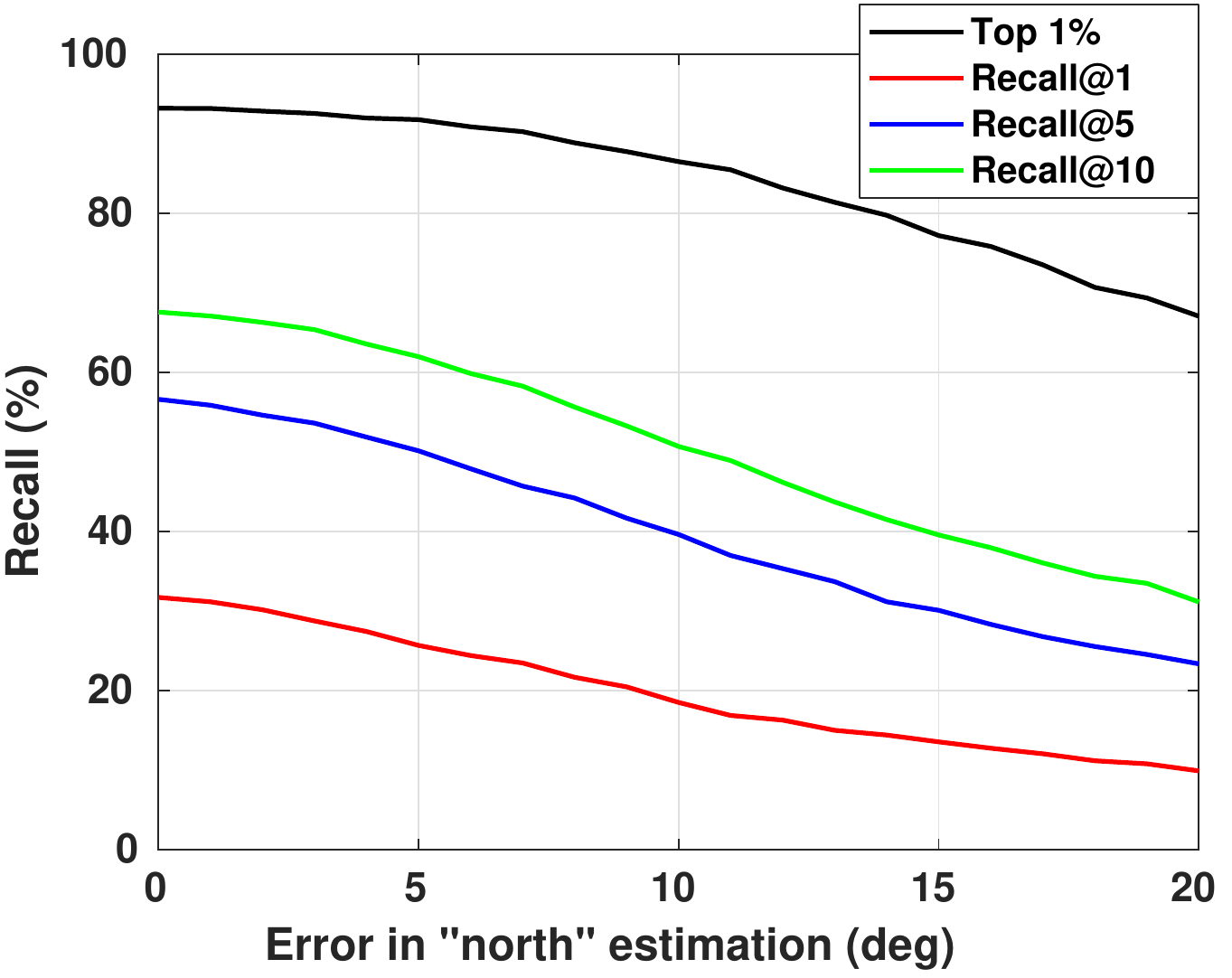}
\end{center}
   \caption{\small Comparison of recalls with respect to errors in the `true north' estimation on CVUSA. Our method degrades gracefully as the error increases.}
\label{fig:northNoise}
\end{figure}

\noindent\textbf{Robustness to errors in orientation estimation.} Our method utilize North-direction aligned street-view panoramas and satellite images for cross-view localization. Note that satellite images are always north-aligned and it is not difficult to roughly align street-view panoramas to the true north with a smart-phone or compass (\eg, a Google Nexus-4 has an average orientation error of $3.6$\degree \cite{ma2013experimental}).  Nevertheless, it is important to know the impact of errors in the estimated `North'. We add different levels of noise between 0 to 20\degree. At each error level, we generate a random angle, and rotate the ground-level panorama by this random angle.  For an equirectangular rectified panorama, this is done by a simple circular crop-and-paste. Figure \ref{fig:northNoise} gives the recall performance at top $1\%$ and recall@K accuracy with different errors. As can be seen, both the recall at top $1\%$ and recall@K decrease gracefully with the increase of error levels.

\begin{figure*}[!ht] 
\centering
\begin{subfigure}[t]{\textwidth}
\centering
\begin{subfigure}[t]{0.26\textwidth}
\includegraphics[width=\textwidth]{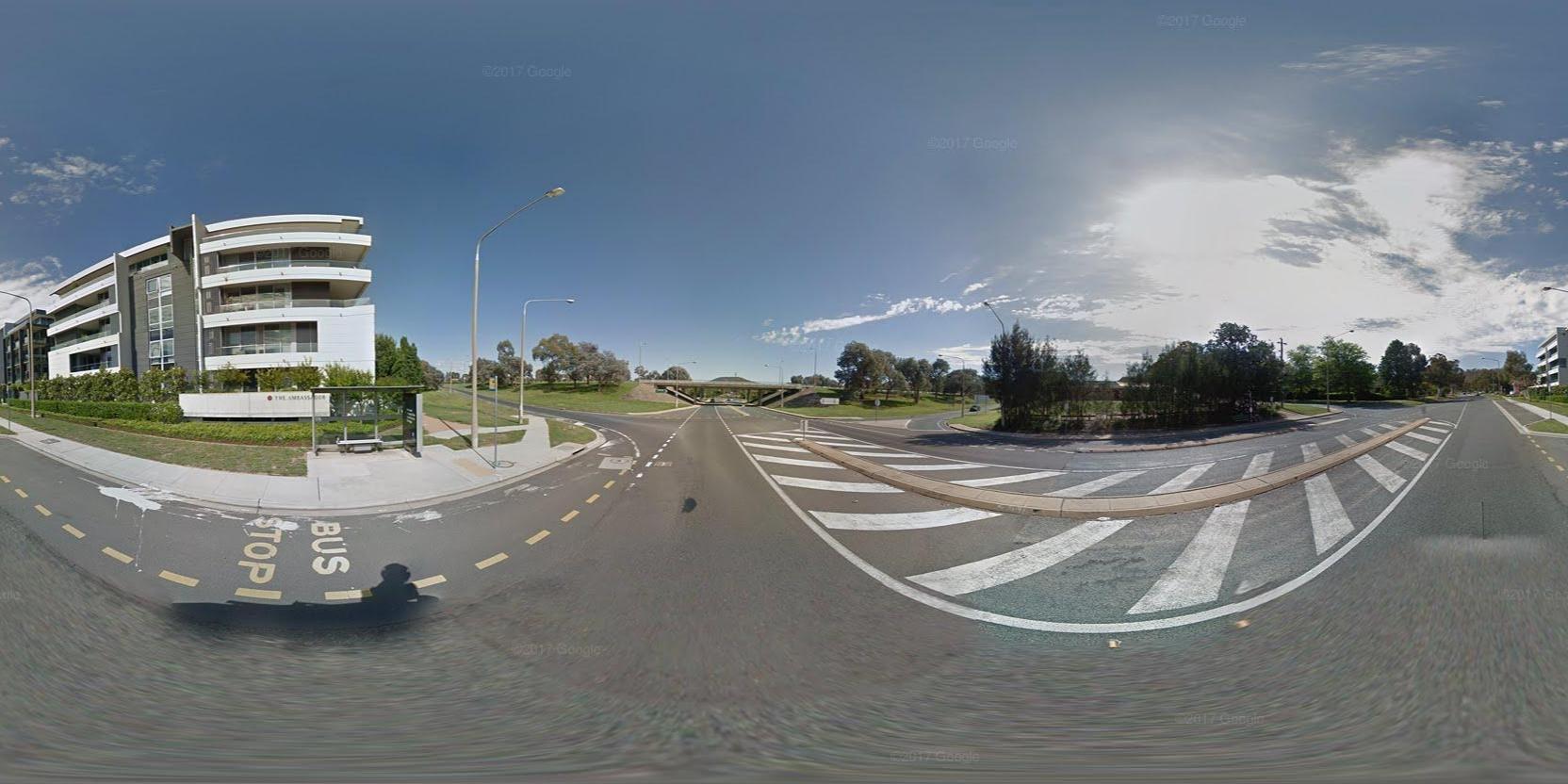}
\end{subfigure}~
\begin{subfigure}[t]{0.13\textwidth}
\includegraphics[width=\textwidth]{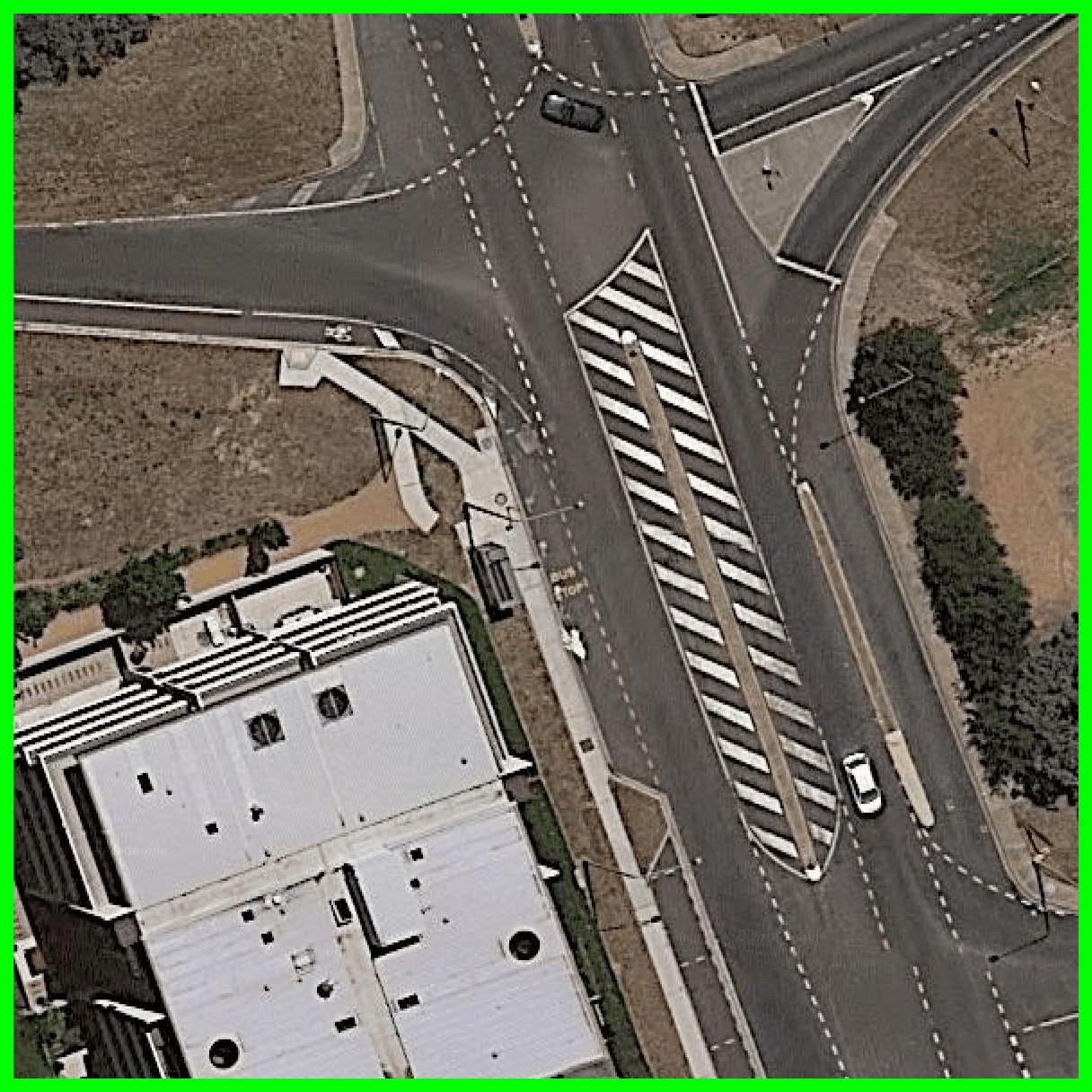}
\end{subfigure}
\begin{subfigure}[t]{0.13\textwidth}
\includegraphics[width=\textwidth]{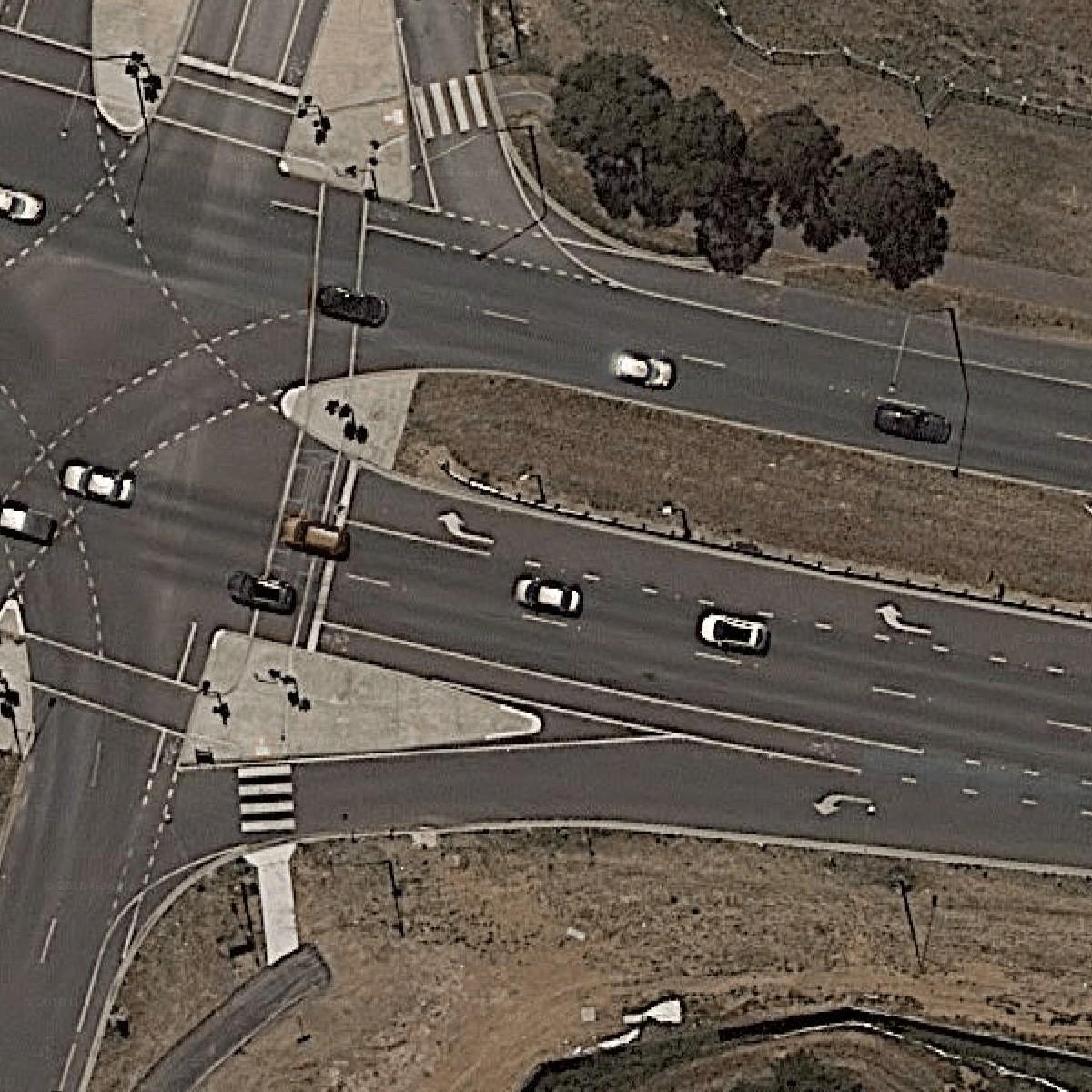}
\end{subfigure}
\begin{subfigure}[t]{0.13\textwidth}
\includegraphics[width=\textwidth]{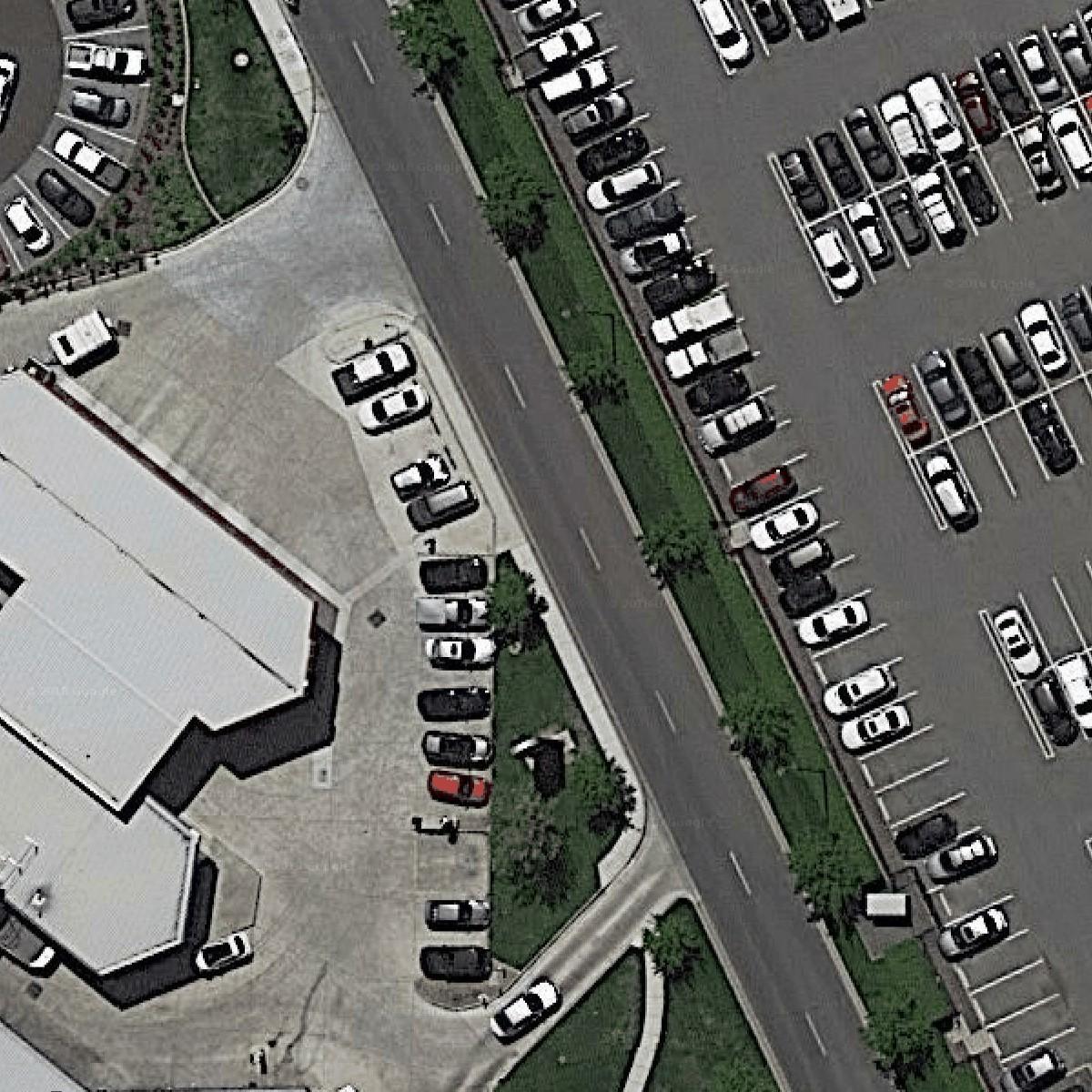}
\end{subfigure}
\begin{subfigure}[t]{0.13\textwidth}
\includegraphics[width=\textwidth]{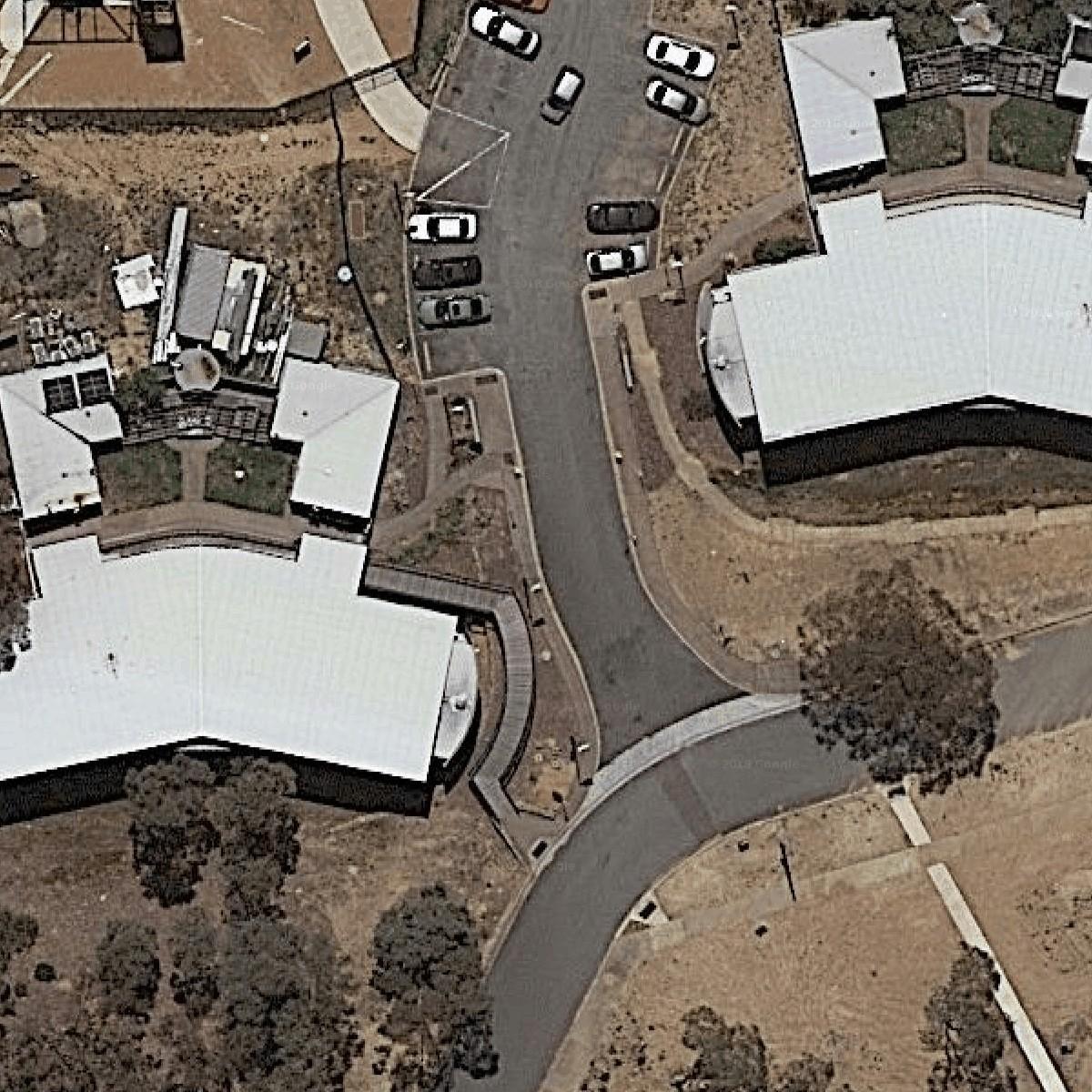}
\end{subfigure}
\begin{subfigure}[t]{0.13\textwidth}
\includegraphics[width=\textwidth]{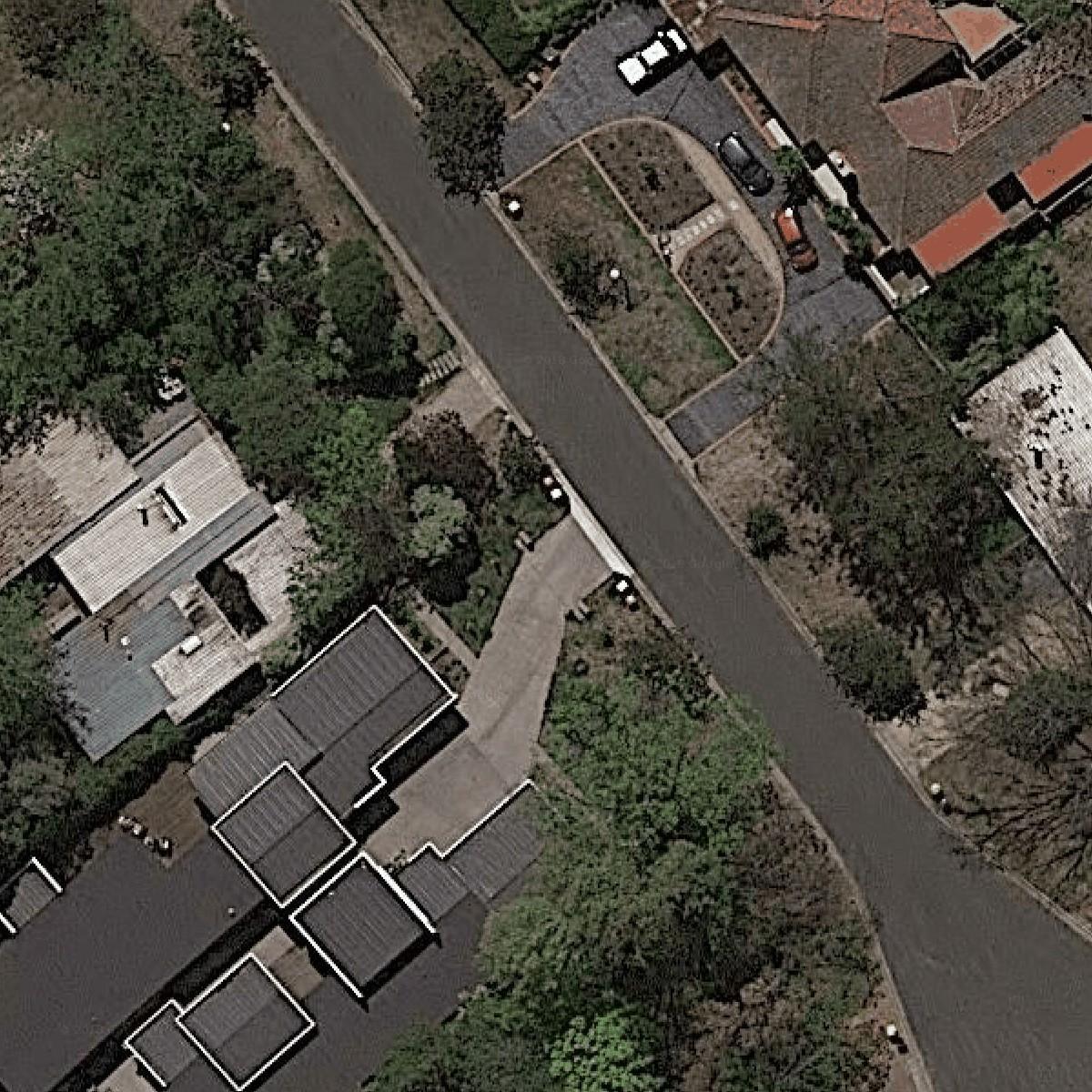}
\end{subfigure}
\end{subfigure}
\vspace{0.02in}
\begin{subfigure}[t]{\textwidth}
\centering
\begin{subfigure}[t]{0.26\textwidth}
\includegraphics[width=\textwidth]{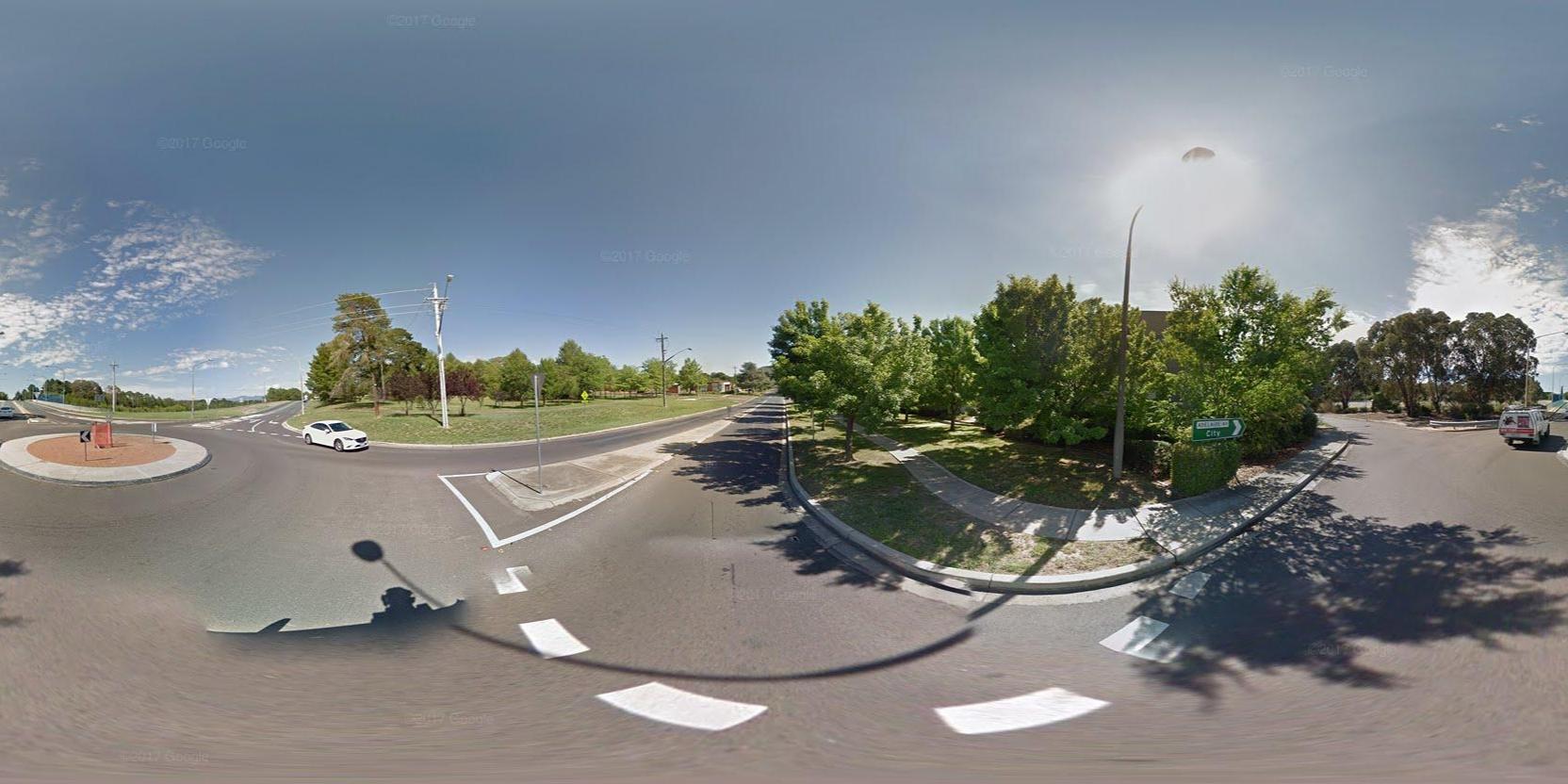}
\end{subfigure}~
\begin{subfigure}[t]{0.13\textwidth}
\includegraphics[width=\textwidth]{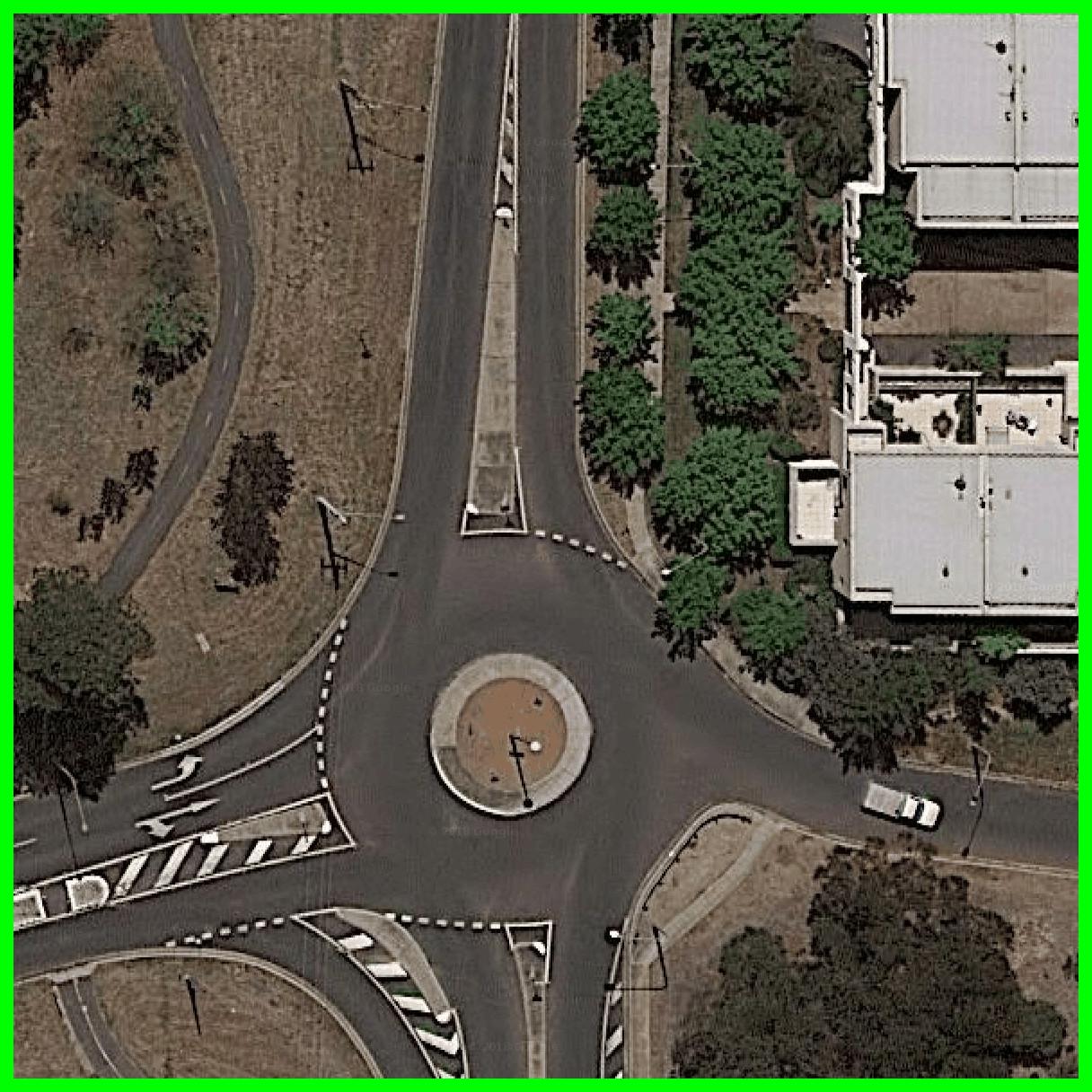}
\end{subfigure}
\begin{subfigure}[t]{0.13\textwidth}
\includegraphics[width=\textwidth]{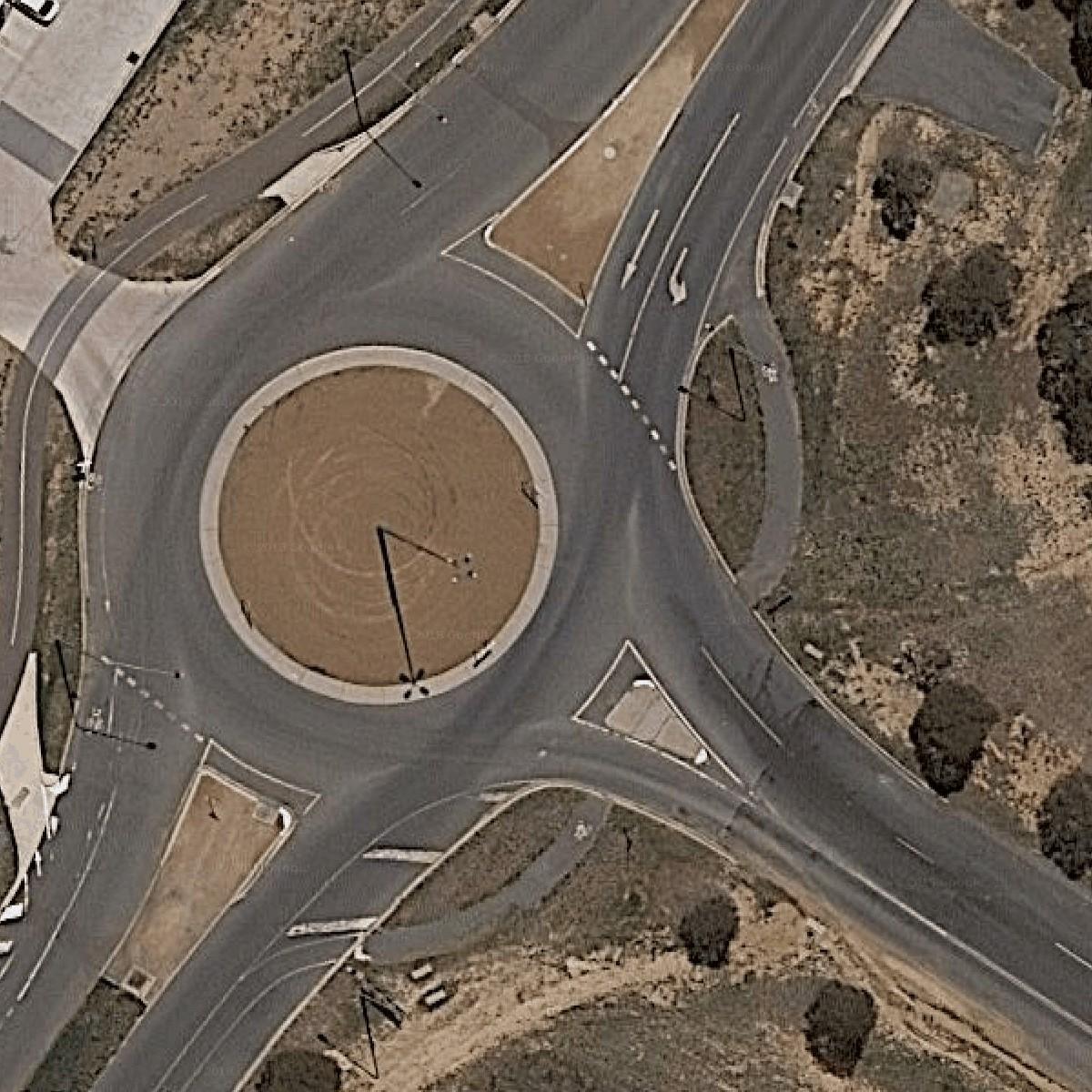}
\end{subfigure}
\begin{subfigure}[t]{0.13\textwidth}
\includegraphics[width=\textwidth]{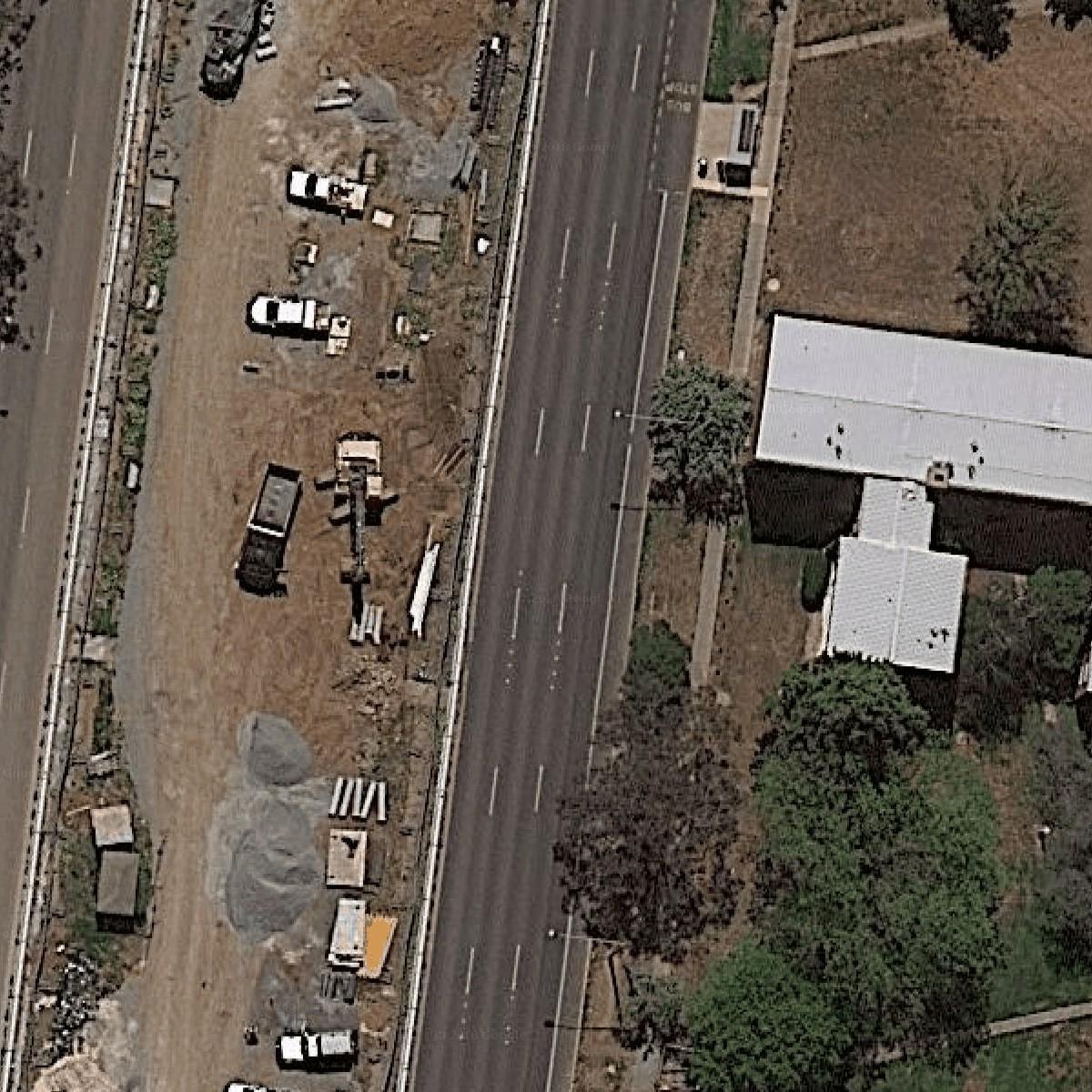}
\end{subfigure}
\begin{subfigure}[t]{0.13\textwidth}
\includegraphics[width=\textwidth]{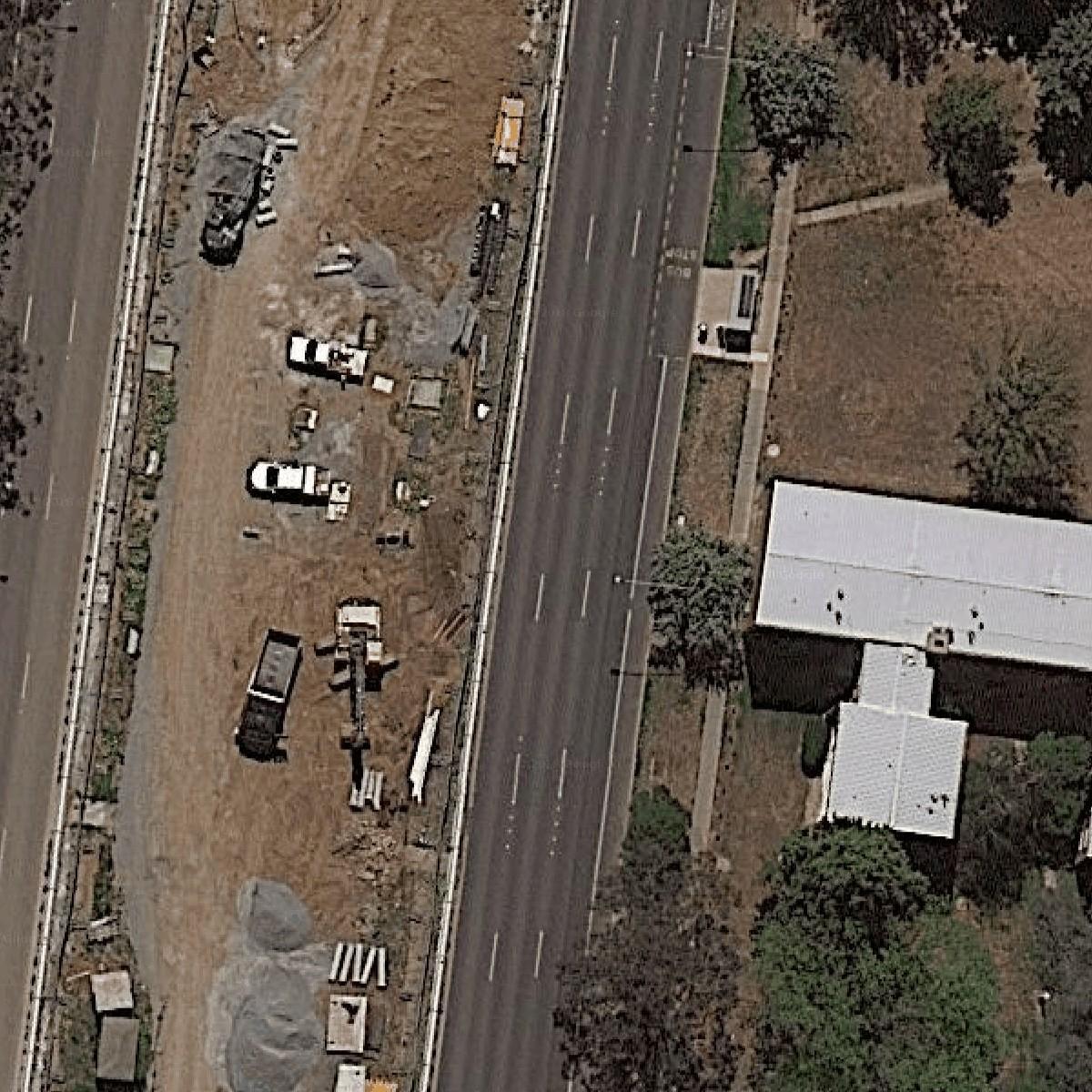}
\end{subfigure}
\begin{subfigure}[t]{0.13\textwidth}
\includegraphics[width=\textwidth]{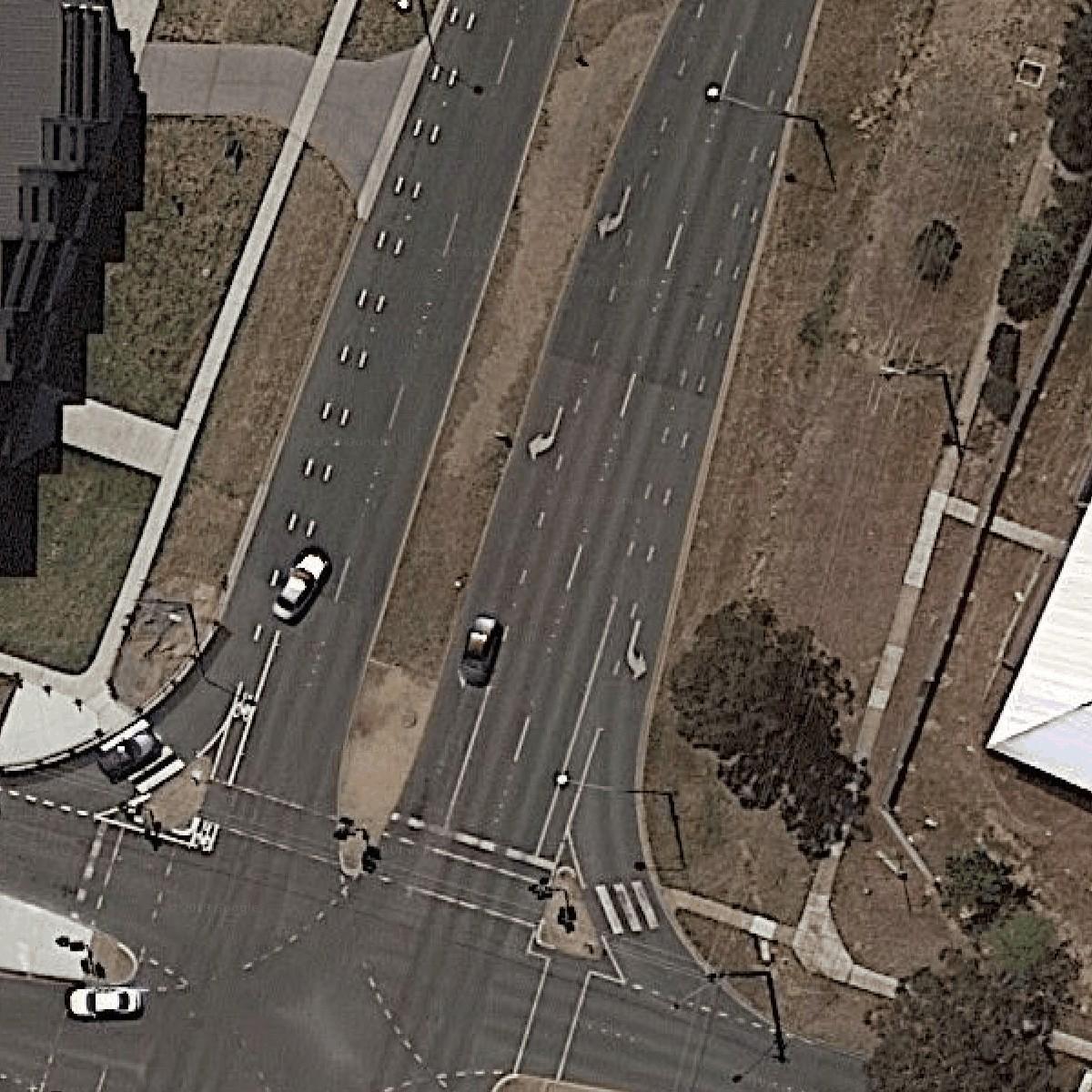}
\end{subfigure}
\end{subfigure}
\begin{subfigure}[t]{\textwidth}
\centering
\begin{subfigure}[t]{0.26\textwidth}
\includegraphics[width=\textwidth]{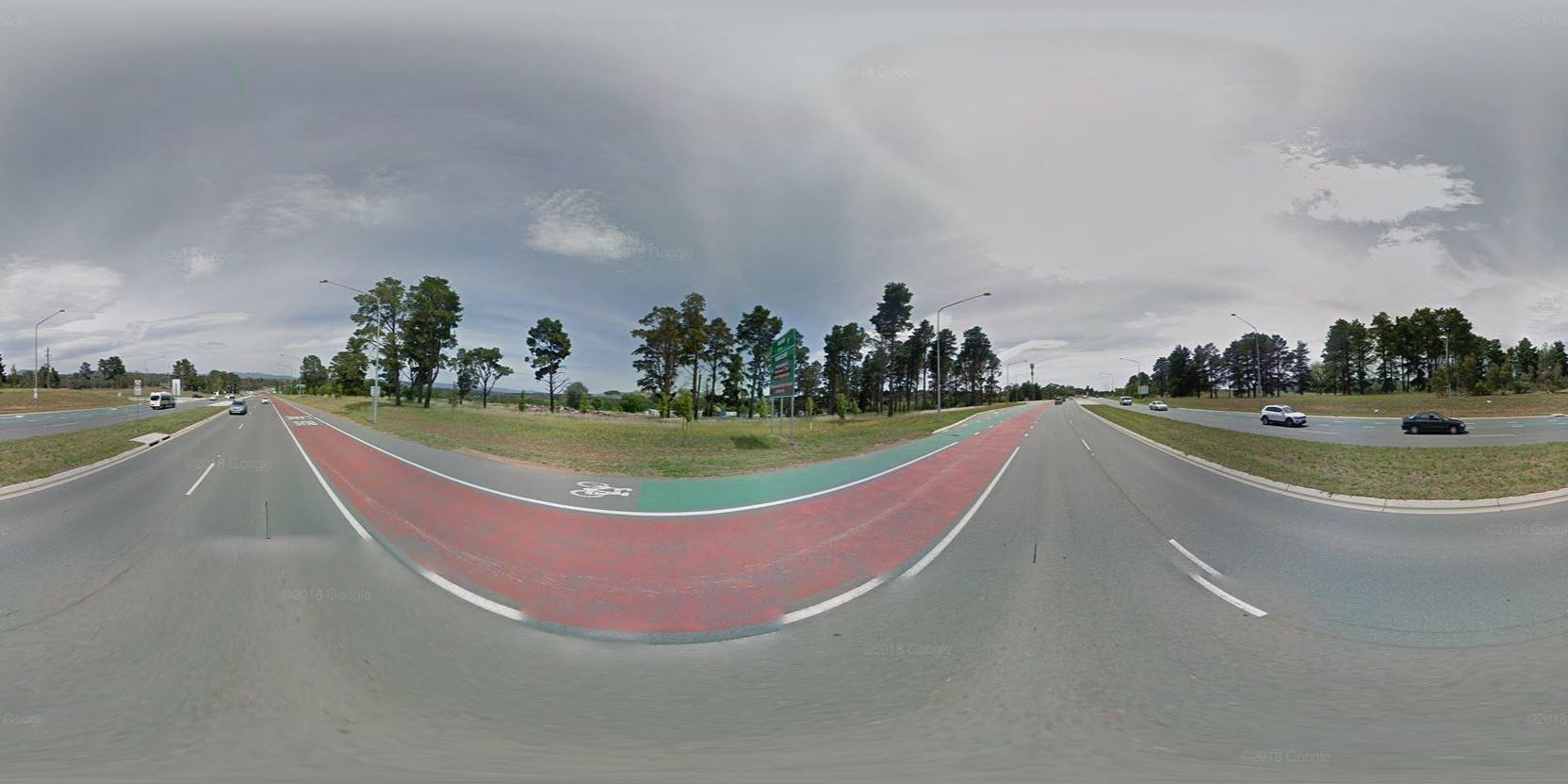}
{\caption{ Query}}
\end{subfigure}~
\begin{subfigure}[t]{0.13\textwidth}
\includegraphics[width=\textwidth]{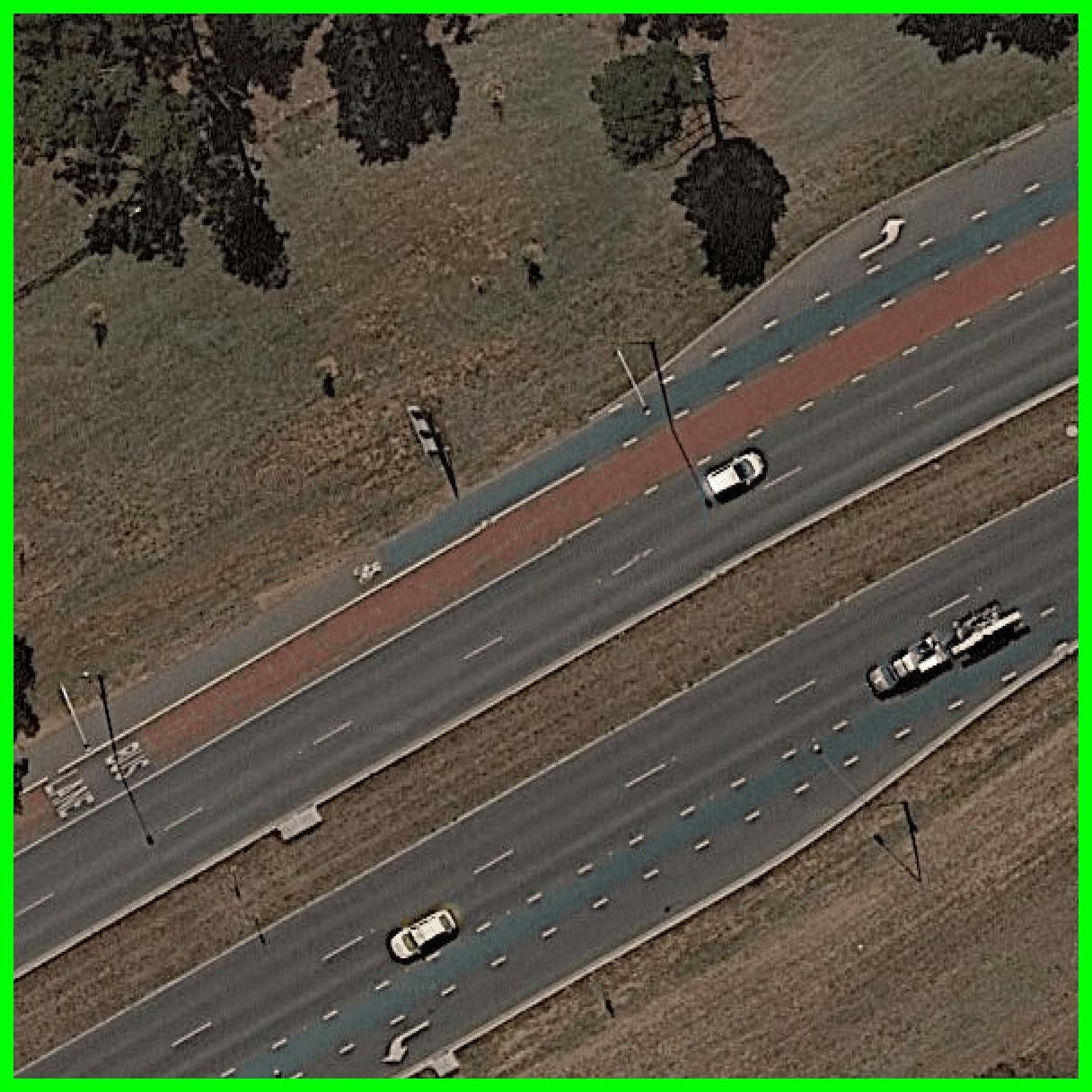}
{\caption{ Top 1}}
\end{subfigure}
\begin{subfigure}[t]{0.13\textwidth}
\includegraphics[width=\textwidth]{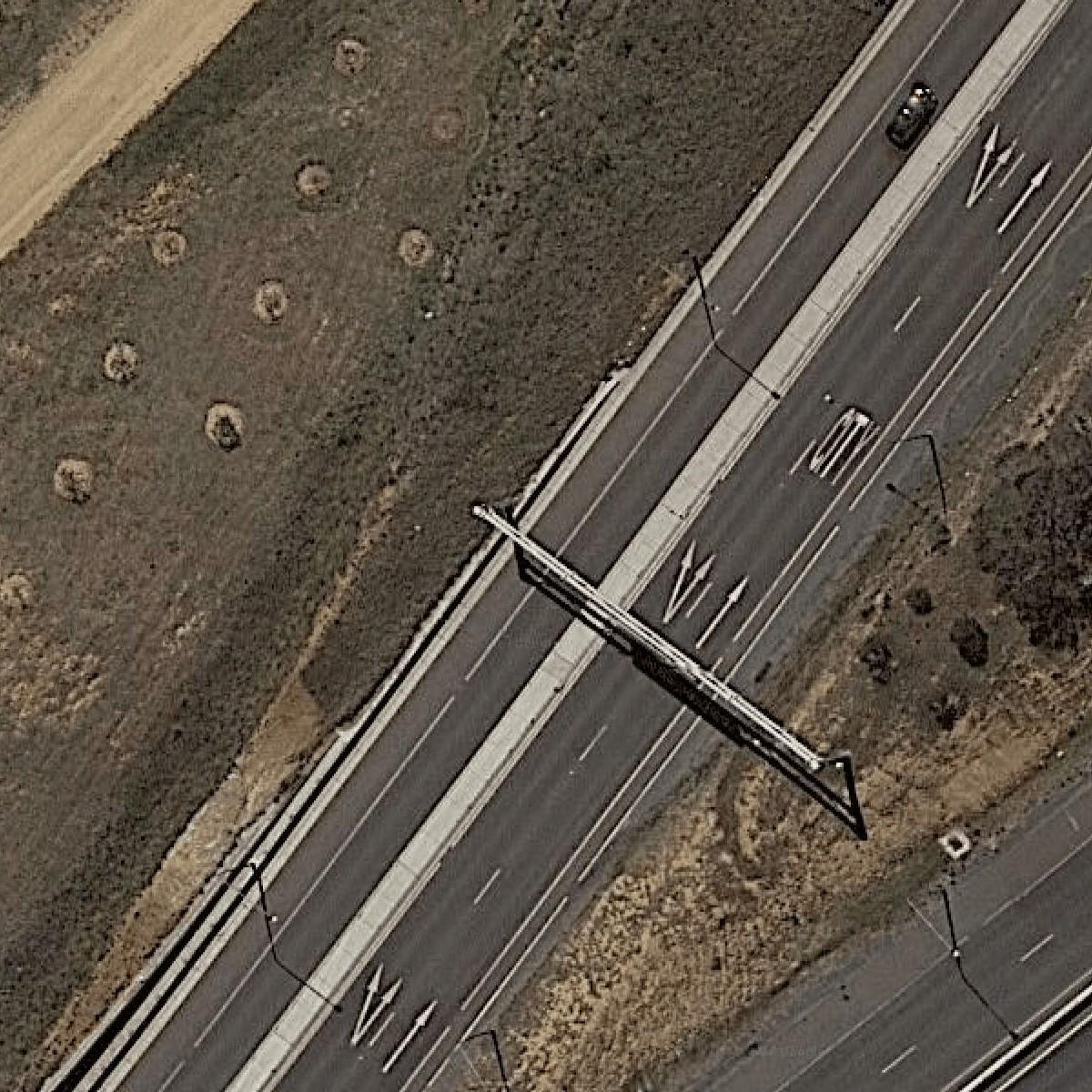}
{\caption{ Top 2}}
\end{subfigure}
\begin{subfigure}[t]{0.13\textwidth}
\includegraphics[width=\textwidth]{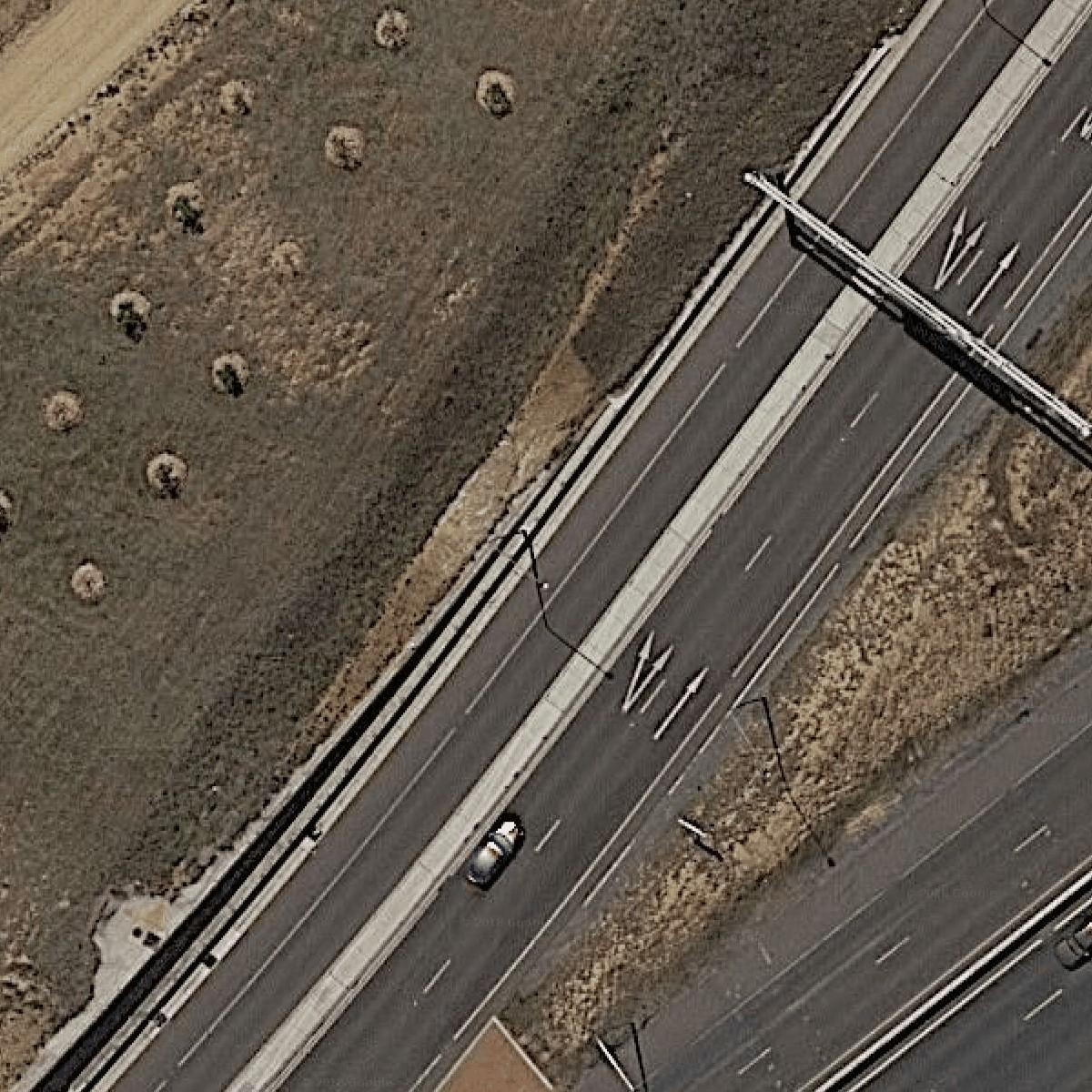}
{\caption{ Top 3}}
\end{subfigure}
\begin{subfigure}[t]{0.13\textwidth}
\includegraphics[width=\textwidth]{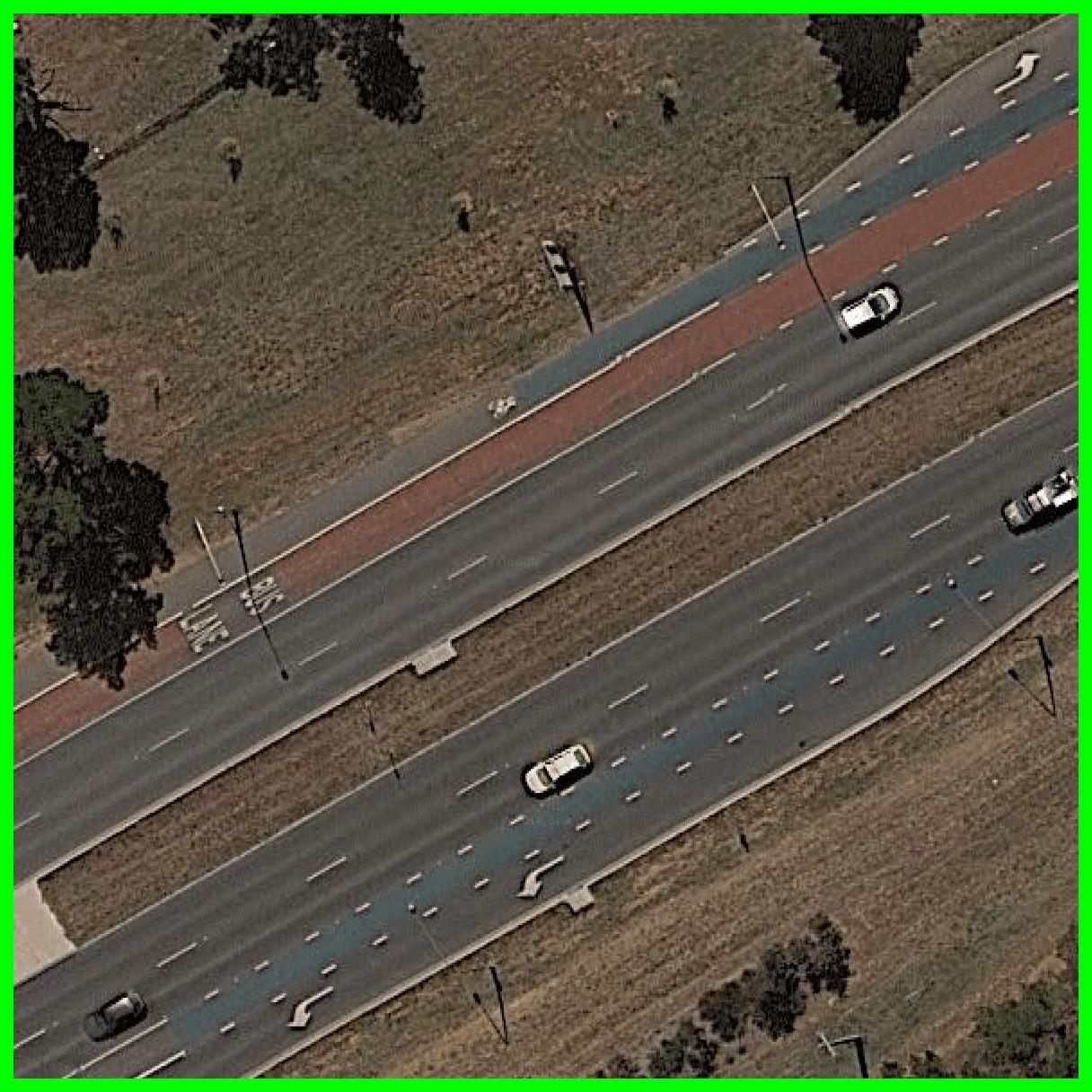}
{\caption{ Top 4}}
\end{subfigure}
\begin{subfigure}[t]{0.13\textwidth}
\includegraphics[width=\textwidth]{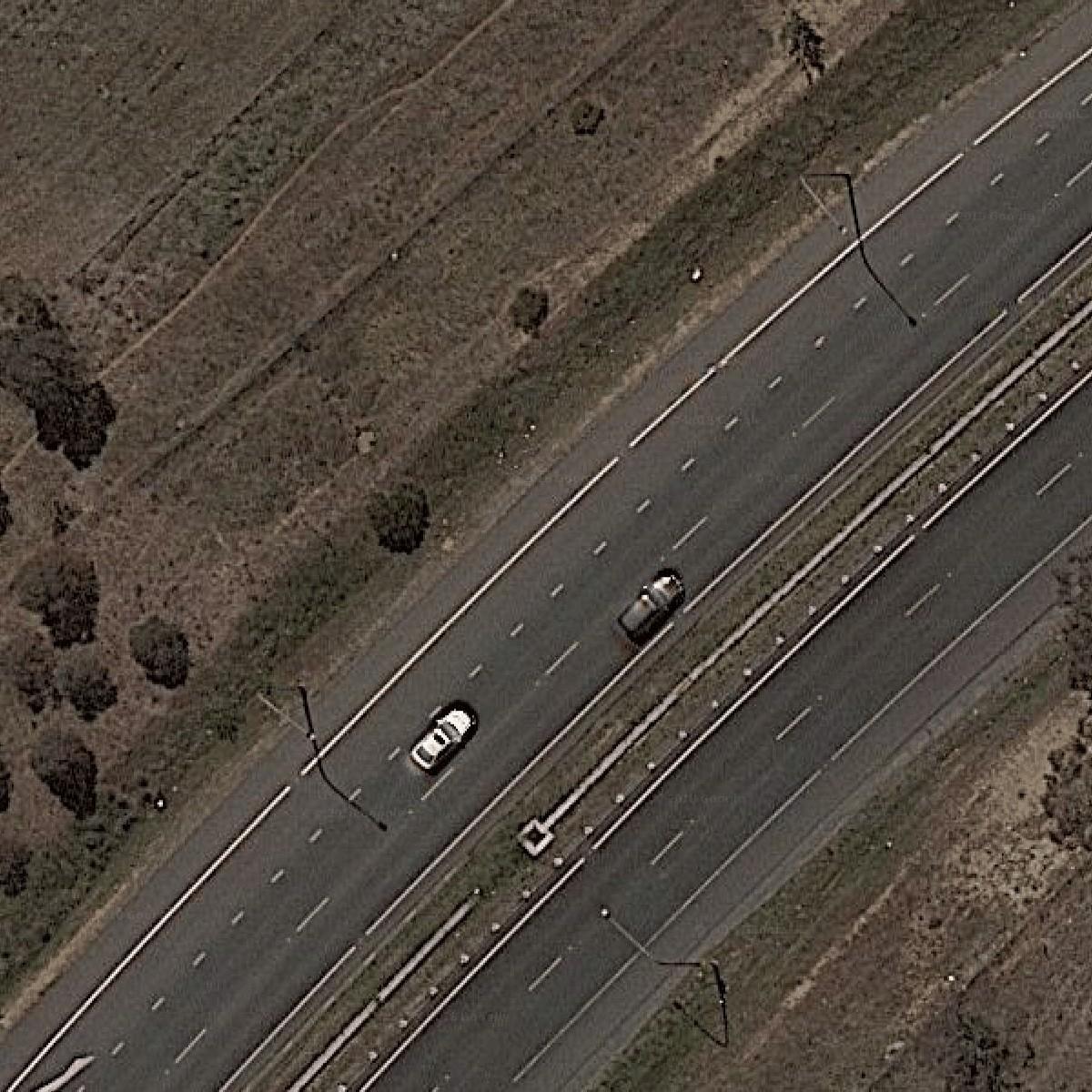}
{\caption{ Top 5}}
\end{subfigure}
\end{subfigure}
\caption{\small{Example localization results on {\bf CVACT dataset} by our method. From left to right: query image and the Top 1-5 retrieved images. Green borders indicate correct retrieved results. Since our dataset densely covers a city-scale environment, a query image may have multiple correct matches (\eg, the $3^{rd}$ row). (Best viewed in color on screen)}}
\label{fig:imageRetrieval}
\end{figure*}



\subsection{ACT city-scale cross-view dataset}\label{Sec::ACTdataset}
To validate the generalization ability of our method on larger-scale geographical localization instances,  we collect and create a new city-scale and fully gps-tagged cross-view dataset (named the `CVACT dataset') densely covering Canberra.  GPS footprints of the dataset are displayed in Figure \ref{fig:sampleImagePairANUdata} (c). 
Street-view panoramas are collected from Google Street View API \cite{GoogleStreetAPI} covering a geographic area of 300 square miles at zoom level $2$. The image resolution of panoramas is $1664\times 832$. Satellite images are collected from Google Map API \cite{GoogleSatAPI}. For each panorama, we download the matchable satellite image at the GPS position of the panorama at the best zoom level $20$. The image resolution of satellite images is $1200\times 1200$ after removing the Google watermark. The ground resolution for satellite images is $0.12$ meters per pixel. A comparison between our CVACT dataset and CVUSA is given in Table-\ref{dataset_compare}. Figure \ref{fig:sampleImagePairANUdata} (b,d) gives a sample cross-view image pair of our dataset.

\begin{table}[!h]
\small{
\caption{Comparison of CVUSA and CVACT datasets}\label{dataset_compare}
\setlength{\tabcolsep}{2.5pt}
\begin{tabular}{|l|l|r|r|r|r|}
\hline
     & \begin{tabular}[c]{@{}l@{}}Ground-view \\ FoV/image res.\end{tabular} &GPS-tag&\begin{tabular}[c]{@{}l@{}}Satellite \\ resolution\end{tabular} & \#training & \#testing \\ \hline
CVACT  & 360/1664x832                                                                & Yes     & 1200x1200                                                      & 35,532      & 92,802     \\ \hline
CVUSA & 360/1232x224                                                                & No      & 750x750                                                        & 35,532      & 8,884      \\ \hline
\end{tabular}
}
\end{table}

\begin{figure}[!h]
\begin{center}
\includegraphics[width=0.36\textwidth]{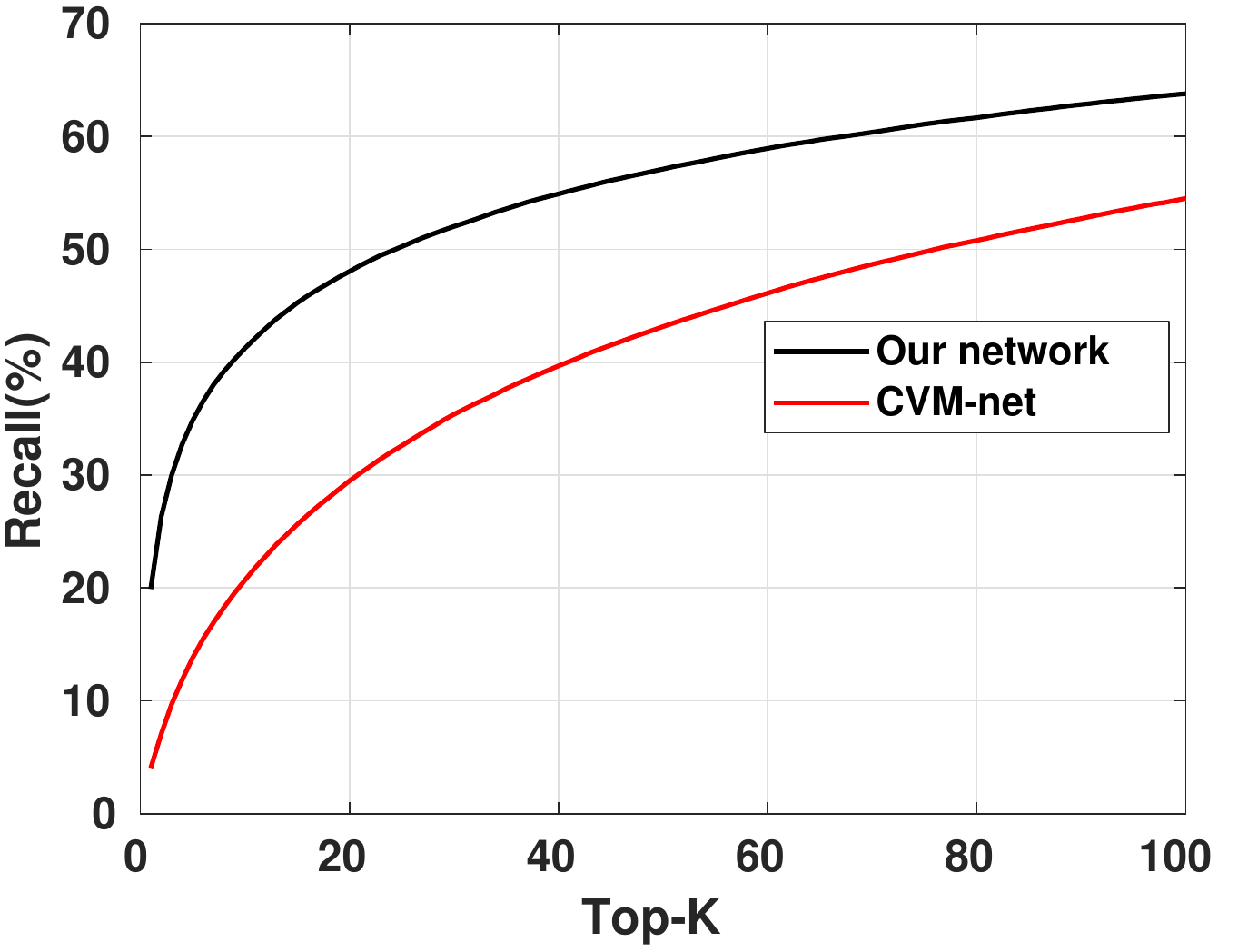}
\end{center} \caption{\small{Localization performance of our method versus CVM-net on our new \textbf{CVACT dataset}.}}
\label{fig:recall_geoLoc}
\end{figure}

Since our dataset is equipped with accurate GPS-tags, we are able to evaluate metric location accuracy.  We tested $92,802$ cross-view image pairs -- viz. $10\times$ bigger than CVUSA dataset \cite{zhai2017predicting}.  We make sure that image pairs in the training set and the testing set are non-overlapping. We use the metric in \cite{arandjelovic2016netvlad} to measure the localization performance. Specifically, the query
image is deemed localized if at least one of the top
$N$ retrieved database satellite images is within $5$ meters from the ground-truth location. A recall@K curve is given in Figure \ref{fig:recall_geoLoc}. We can see that our method outperforms \cite{Hu_2018_CVPR}, with an improvement of $15.84\%$ at top-1. This result also reveals the difficulty of our new dataset, namely only $19.90\%$ query images get to be localized within an accuracy of $\le5m$-level; We hope this will motivate other researchers to tackle this challenge task and use our CVACT dataset.  Some example localization results by our method are shown in Figure \ref{fig:imageRetrieval}.

\section{Conclusion}

Image-based geo-localization is essentially  a {\em geometry} problem, where the ultimate goal is to recover 6-DoF camera pose (\ie, both location and orientation). As such, applying geometric cues  (\eg,  orientation)  to localization is both natural and desirable.  However, most previous image-based localization methods have either overlooked such important cues,  or  have no effective way to incorporate such geometry information.   Instead, they treat the problem as a pure content-based image retrieval task, and focus on finding visual similarity in terms of appearance and semantic contents of the images. 

In this work,  we have successfully demonstrated that, by adding a simple orientation map we are able to teach a Siamese localization network the (geometric) notion of orientation.   This results in significant  improvement in localization performance (\eg,  our top 1\% recall rate is boosted by over 25\% compared with without using orientations).   Our method for adding orientation map to a neural network is simple, and transparent;  the same idea may be applied to other types of deep networks or for different applications as well.    It is our position that, in solving geometry-related vision problems, whenever geometry clues (or insights) are available, one should always consider how to exploit them, rather than training a CNN end-to-end as a blind black-box.   We hope our idea can inspire other researchers working on related problems.   Our second contribution of this paper is a large-scale,  fully-annotated and geo-referenced cross-view image localization dataset -- the CVACT dataset.  We hope it is a valuable  addition to the localization benchmark and literature.   

\section*{Acknowledgement}

This research was supported in part by the Australian Research Council (ARC) grants (CE140100016) and Australia Centre for Robotic Vision. Hongdong Li is also funded in part by ARC-DP (190102261) and ARC-LE (190100080). We
gratefully acknowledge the support of NVIDIA Corporation with the donation of the GPU. We thank all anonymous reviewers for their valuable comments.

\newpage
{\small
\bibliographystyle{ieee}
\bibliography{egbib}

\begin{thebibliography}{10}\itemsep=-1pt

\bibitem{Animal_navigation}
Animal navigation.
\newblock \url{https://en.wikipedia.org/wiki/Animal_navigation}.

\bibitem{GoogleSatAPI}
Google satellite image api.
\newblock
  \url{https://developers.google.com/maps/documentation/maps-static/intro}.

\bibitem{GoogleStreetAPI}
Google street view api.
\newblock
  \url{https://developers.google.com/maps/documentation/streetview/intro}.

\bibitem{Toolbox_optical}
A toolbox to visualize dense image correspondences.
\newblock \url{https://hci.iwr.uni-heidelberg.de/Correspondence_Visualization}.

\bibitem{arandjelovic2016netvlad}
Relja Arandjelovic, Petr Gronat, Akihiko Torii, Tomas Pajdla, and Josef Sivic.
\newblock Netvlad: Cnn architecture for weakly supervised place recognition.
\newblock In {\em Proceedings of the IEEE Conference on Computer Vision and
  Pattern Recognition}, pages 5297--5307, 2016.

\bibitem{castaldo2015semantic}
Francesco Castaldo, Amir Zamir, Roland Angst, Francesco Palmieri, and Silvio
  Savarese.
\newblock Semantic cross-view matching.
\newblock In {\em Proceedings of the IEEE International Conference on Computer
  Vision Workshops}, pages 9--17, 2015.

\bibitem{deng2009imagenet}
Jia Deng, Wei Dong, Richard Socher, Li-Jia Li, Kai Li, and Li Fei-Fei.
\newblock Imagenet: A large-scale hierarchical image database.
\newblock In {\em Computer Vision and Pattern Recognition, 2009. CVPR 2009.
  IEEE Conference on}, pages 248--255. Ieee, 2009.

\bibitem{dollar2009integral}
Piotr Doll{\'a}r, Zhuowen Tu, Pietro Perona, and Serge Belongie.
\newblock Integral channel features.
\newblock 2009.

\bibitem{guberman2016complex}
Nitzan Guberman.
\newblock On complex valued convolutional neural networks.
\newblock {\em arXiv preprint arXiv:1602.09046}, 2016.

\bibitem{he2016deep}
Kaiming He, Xiangyu Zhang, Shaoqing Ren, and Jian Sun.
\newblock Deep residual learning for image recognition.
\newblock In {\em Proceedings of the IEEE conference on computer vision and
  pattern recognition}, pages 770--778, 2016.

\bibitem{hu2018cvm}
Sixing Hu, Mengdan Feng, Rang~MH Nguyen, and G~Hee Lee.
\newblock Cvm-net: Cross-view matching network for image-based ground-to-aerial
  geo-localization.
\newblock In {\em IEEE Conference on Computer Vision and Pattern Recognition
  (CVPR)}, 2018.

\bibitem{Hu_2018_CVPR}
Sixing Hu, Mengdan Feng, Rang M.~H. Nguyen, and Gim~Hee Lee.
\newblock Cvm-net: Cross-view matching network for image-based ground-to-aerial
  geo-localization.
\newblock In {\em The IEEE Conference on Computer Vision and Pattern
  Recognition (CVPR)}, June 2018.

\bibitem{huang2017densely}
Gao Huang, Zhuang Liu, Laurens Van Der~Maaten, and Kilian~Q Weinberger.
\newblock Densely connected convolutional networks.
\newblock In {\em CVPR}, volume~1, page~3, 2017.

\bibitem{isola2017image}
Phillip Isola, Jun-Yan Zhu, Tinghui Zhou, and Alexei~A Efros.
\newblock Image-to-image translation with conditional adversarial networks.
\newblock In {\em Proceedings of the IEEE conference on computer vision and
  pattern recognition}, pages 1125--1134, 2017.

\bibitem{kingma2014adam}
Diederik~P Kingma and Jimmy Ba.
\newblock Adam: A method for stochastic optimization.
\newblock {\em arXiv preprint arXiv:1412.6980}, 2014.

\bibitem{lin2013cross}
Tsung-Yi Lin, Serge Belongie, and James Hays.
\newblock Cross-view image geolocalization.
\newblock In {\em Proceedings of the IEEE Conference on Computer Vision and
  Pattern Recognition}, pages 891--898, 2013.

\bibitem{lin2015learning}
Tsung-Yi Lin, Yin Cui, Serge Belongie, and James Hays.
\newblock Learning deep representations for ground-to-aerial geolocalization.
\newblock In {\em Proceedings of the IEEE conference on computer vision and
  pattern recognition}, pages 5007--5015, 2015.

\bibitem{Liu_2017_ICCV}
Liu Liu, Hongdong Li, and Yuchao Dai.
\newblock Efficient global 2d-3d matching for camera localization in a
  large-scale 3d map.
\newblock In {\em The IEEE International Conference on Computer Vision (ICCV)},
  Oct 2017.

\bibitem{DBLP:journals/corr/abs-1808-08779}
Liu Liu, Hongdong Li, and Yuchao Dai.
\newblock Deep stochastic attraction and repulsion embedding for image based
  localization.
\newblock {\em CoRR}, abs/1808.08779, 2018.

\bibitem{liu2018robust}
Liu Liu, Hongdong Li, Yuchao Dai, and Quan Pan.
\newblock Robust and efficient relative pose with a multi-camera system for
  autonomous driving in highly dynamic environments.
\newblock {\em IEEE Transactions on Intelligent Transportation Systems},
  19(8):2432--2444, 2018.

\bibitem{ma2013experimental}
Zhizhong Ma, Yuansong Qiao, Brian Lee, and Enda Fallon.
\newblock Experimental evaluation of mobile phone sensors.
\newblock 2013.

\bibitem{maaten2008visualizing}
Laurens van~der Maaten and Geoffrey Hinton.
\newblock Visualizing data using t-sne.
\newblock {\em Journal of machine learning research}, 9(Nov):2579--2605, 2008.

\bibitem{DBLP:journals/corr/MousavianK16}
Arsalan Mousavian and Jana Kosecka.
\newblock Semantic image based geolocation given a map.
\newblock {\em CoRR}, abs/1609.00278, 2016.

\bibitem{radenovic2018fine}
Filip Radenovi{\'c}, Giorgos Tolias, and Ondrej Chum.
\newblock Fine-tuning cnn image retrieval with no human annotation.
\newblock {\em IEEE Transactions on Pattern Analysis and Machine Intelligence},
  2018.

\bibitem{regmi2018cross}
Krishna Regmi and Ali Borji.
\newblock Cross-view image synthesis using conditional gans.
\newblock In {\em Proceedings of the IEEE Conference on Computer Vision and
  Pattern Recognition}, pages 3501--3510, 2018.

\bibitem{ronneberger2015u}
Olaf Ronneberger, Philipp Fischer, and Thomas Brox.
\newblock U-net: Convolutional networks for biomedical image segmentation.
\newblock In {\em International Conference on Medical image computing and
  computer-assisted intervention}, pages 234--241. Springer, 2015.

\bibitem{sattler2019understanding}
Torsten Sattler, Qunjie Zhou, Marc Pollefeys, and Laura Leal-Taixe.
\newblock Understanding the limitations of cnn-based absolute camera pose
  regression.
\newblock {\em arXiv preprint arXiv:1903.07504}, 2019.

\bibitem{simonyan2014very}
Karen Simonyan and Andrew Zisserman.
\newblock Very deep convolutional networks for large-scale image recognition.
\newblock {\em arXiv preprint arXiv:1409.1556}, 2014.

\bibitem{tian2017cross}
Yicong Tian, Chen Chen, and Mubarak Shah.
\newblock Cross-view image matching for geo-localization in urban environments.
\newblock In {\em IEEE Conference on Computer Vision and Pattern Recognition
  (CVPR)}, pages 1998--2006, 2017.

\bibitem{trabelsi2018deep}
Chiheb Trabelsi, Olexa Bilaniuk, Ying Zhang, Dmitriy Serdyuk, Sandeep
  Subramanian, Joao~Felipe Santos, Soroush Mehri, Negar Rostamzadeh, Yoshua
  Bengio, and Christopher~J Pal.
\newblock Deep complex networks.
\newblock In {\em International Conference on Learning Representations}, 2018.

\bibitem{vo2016localizing}
Nam~N Vo and James Hays.
\newblock Localizing and orienting street views using overhead imagery.
\newblock In {\em European Conference on Computer Vision}, pages 494--509.
  Springer, 2016.

\bibitem{Wang_2017_ICCV}
Jian Wang, Feng Zhou, Shilei Wen, Xiao Liu, and Yuanqing Lin.
\newblock Deep metric learning with angular loss.
\newblock In {\em The IEEE International Conference on Computer Vision (ICCV)},
  Oct 2017.

\bibitem{workman2015location}
Scott Workman and Nathan Jacobs.
\newblock On the location dependence of convolutional neural network features.
\newblock In {\em Proceedings of the IEEE Conference on Computer Vision and
  Pattern Recognition Workshops}, pages 70--78, 2015.

\bibitem{workman2015wide}
Scott Workman, Richard Souvenir, and Nathan Jacobs.
\newblock Wide-area image geolocalization with aerial reference imagery.
\newblock In {\em Proceedings of the IEEE International Conference on Computer
  Vision}, pages 3961--3969, 2015.

\bibitem{zemel1995lending}
Richard~S Zemel, Christopher~KI Williams, and Michael~C Mozer.
\newblock Lending direction to neural networks.
\newblock {\em Neural Networks}, 8(4):503--512, 1995.

\bibitem{zhai2017predicting}
Menghua Zhai, Zachary Bessinger, Scott Workman, and Nathan Jacobs.
\newblock Predicting ground-level scene layout from aerial imagery.
\newblock In {\em IEEE Conference on Computer Vision and Pattern Recognition},
  volume~3, 2017.

\bibitem{zhou2014learning}
Bolei Zhou, Agata Lapedriza, Jianxiong Xiao, Antonio Torralba, and Aude Oliva.
\newblock Learning deep features for scene recognition using places database.
\newblock In {\em Advances in neural information processing systems}, pages
  487--495, 2014.

\end{thebibliography}
}

\end{document}